# A model of rational interlocutors: Unification of comprehension and production

Hanlin Wu[1] and Zhenguang G. Cai[1,2]
[1]Department of Linguistics and Modern Languages, The Chinese University of Hong Kong, Sha Tin, N.T., Hong Kong
[2]Brain and Mind Institute, The Chinese University of Hong Kong, Sha Tin, N.T., Hong Kong
Corresponding author: Hanlin Wu, hanlinwu@cuhk.edu.hk

## Abstract

Who we communicate with influences both our interpretation of their utterances and the design of our own. Such adjustment to the conversational partner is studied as speaker modeling in comprehension and as audience design in production, with the two literatures having developed largely separately. We argue that both adjustments express one rational computation and propose the rational interlocutor (RI) model, a computational account unifying comprehension and production. An interlocutor maintains a model of their partner, defined by three parameters: an identity parameter $\Pi$ sets the messages and forms expected from the partner; a fidelity parameter $\Phi$ sets how reliably messages and utterances map onto each other for them; a knowledge parameter $\Lambda$ sets how knowledgeable the partner is believed to be. Comprehension and production are thus mirror-image modes of one computation over the partner model. Comprehension chooses the message the partner most likely intends to convey, weighing how well each candidate fits the utterance against how likely this partner is to mean it. Production chooses the utterance from which the partner will best recover the message, weighed against the effort of saying it. This explains why comprehenders appear to rely less on the forms produced by a linguistically less competent speaker, while producers tend to invest more effort in designing forms for them. We conjecture that perceived linguistic competence decomposes into two of these quantities: fidelity and knowledge. Their contrasting profiles across second-language (L2) adults, children, and artificial partners produce distinct and testable predictions.

## 1. Introduction

Who we are communicating with influences both our comprehension of their utterances and the production of our own. In comprehension, when an adult voice says “I cannot sleep without my *teddy bear* in my arms”, the content of the utterance clashes with what the voice suggests about the speaker. Listeners’ brain responses to such utterances resembled their responses to semantic anomalies, within a few hundred milliseconds after hearing the critical word (Van Berkum et al., 2008). In production, speakers make the complementary adjustment and design utterances for their addressee. For example, people describing postcards of New York scenes used bare place names once their addressee was believed to be a fellow New Yorker, and they used fuller descriptions when their addressee was not (Isaacs & Clark, 1987).

In both comprehension and production, an interlocutor relies on a partner model, a mental representation of the person they are communicating with. Comprehension research has developed this idea as a speaker model that conditions how an utterance is perceived and interpreted (Cai et al., 2017; Wu & Cai, 2026a). Production research has developed it as audience design, the tailoring of an utterance to the addressee’s needs (Bell, 1984; Clark & Murphy, 1982). On this account, the two adjustments draw on the same partner model. In comprehension, the “teddy bear” effect arises because the voice identifies the speaker and that identity sets what content is expected. In production, the postcard adjustments follow because the addressee’s identity sets which description will suffice.

A partner model represents many attributes of the conversational partner. It represents age: listeners expect a child speaker to reuse an established label rather than switch to a synonym (Wu et al., 2024). It represents gender: hearing “The first time I got *pregnant* I had a hard time” in a male voice elicits the error-like brain response (Wu & Cai, 2026b). It represents social background: hearing “I have a large *tattoo* on my back” in an upper-class accent elicits a response similar to semantic anomaly (Van Berkum et al., 2008). It also represents dialect background: the word “bonnet” is interpreted as an engine cover in a British accent but as a hat in an American accent (Cai et al., 2017). Among these attributes, the one that has drawn particular attention is linguistic competence. Unlike the other attributes, it produces opposite adjustments in comprehension and in production, and we set out to explain why.

There is good evidence that people comprehend linguistically less competent people such as second language (L2) users in a different way than they do with native speakers. For example, people hearing an implausible sentence such as “the mother gave the candle the daughter” sometimes recover the non-literal alternative “the mother gave the candle *to* the daughter” (Cai et al., 2022; Gibson et al., 2013). They did so more often when the sentence was spoken with a foreign accent (Gibson et al., 2017). A Dutch grammatical-gender error elicited no reliable neural index of repair from a Turkish-accented speaker of Dutch, although the same error from a native speaker did (Hanulíková et al., 2012). An under-informative statement, such as “*some* people have noses with two nostrils”, was rated as making more sense when it was attributed to a non-native speaker with a strong accent (Fairchild & Papafragou, 2018). A filled pause, likewise, no longer led to anticipatory looks toward the object whose name is harder to retrieve when listeners heard a speaker introduced as non-native (Bosker et al., 2014). In each of these cases, comprehenders *discount* a linguistically less competent speaker’s utterance. They tolerate its errors, set its disfluencies aside, and rely more on their own expectations about what was likely meant.

On the other side of the conversation, there is an opposite adjustment. Rather than discount a linguistically less competent partner, speakers *invest* more in them in production. Speakers reuse a child’s or a non-native adult’s expressions more than they reuse a native adult’s (Cai et al., 2021). They also reuse a computer partner’s words more when a start-up screen presented it as a basic rather than an advanced system (Pearson et al., 2006). They use fuller referring expressions rather than just a pronoun, for younger children more than for older ones and for adult L2 learners more than for native-speaking adults (Tal et al., 2023). They hyperarticulate their vowels for infants and for foreigners more than for native adult addressees (Uther et al., 2007), and they slow their speech for a voice assistant relative to a human addressee in pre-scripted spoken interaction (Cohn & Zellou, 2021) and in imagined scenarios (Cohn, Mengesha, et al., 2024). Across these cases, speakers invest more in a linguistically less competent partner: they expend additional effort in order to be understood.

An apparent *discounting-vs-investing paradox* thus emerges: the same belief that a partner is linguistically less competent seems to make listeners discount the utterance and speakers invest in it. The two sets of findings have developed in largely separate research traditions. Work on comprehension asks how a listener uses their knowledge of a speaker to recover the message, and work on production asks how a speaker tailors utterances to the listener. The two adjustments have nonetheless been proposed to be one mechanism termed interlocutor modeling (Cai et al., 2017, 2021). On that proposal an interlocutor builds a model of their conversational partner and uses it in both comprehension and production, once to determine what the partner means and once to be understood by them. Discounting and investment would then be the same response to the same belief, viewed from the two ends of a conversation.

The current paper develops the interlocutor-modeling framework into the *rational interlocutor (RI) model*, a computational model unifying both comprehension and production. We argue that the adjustments during conversations are what a rational interlocutor does when communicating through an imperfect information channel. In such a channel, a partner's intended message and the utterance that carries it can dissociate. The conversational partner enters the model as three parameters: an identity parameter $\Pi$ sets what the partner is expected to mean, the forms they are expected to use, and what is appropriate to say to them; a fidelity parameter $\Phi$ sets how reliably messages and utterances map onto each other for them; a knowledge parameter $\Lambda$ sets how knowledgeable the partner is believed to be.

This paper is organized as follows. We first review the evidence of interlocutor adjustments in comprehension (Section 2) and in production (Section 3). We then propose the RI model (Section 4), which gives the three parameters definite roles: the identity sets the typical messages and forms; the fidelity is the sharpness of the channel linking messages and utterances; and the knowledge weights the message prior. We conjecture that the single scale of linguistic competence should be decomposed into two of the three, fidelity and knowledge. In Section 5 we turn to acquisition of the partner model, the conditions of its use, and differences among interlocutors in how far they rely on it. In Section 6 we relate the RI model to previous accounts and state the predictions that would test it, and we conclude in Section 7.

## 2. Comprehension

We begin with comprehension and ask how a listener's interpretation of the same utterance shifts with who is believed to be speaking. We organize the evidence by level of analysis and identify two main patterns. Listeners rely less on the utterances produced by a linguistically less competent speaker and more on their own expectations. The speaker's identity also sets what that speaker is expected to mean and which forms they are expected to use.

## 2.1 Sentence repair

Sentence interpretation supplies direct evidence that listeners discount a linguistically less competent speaker's utterance. Gibson et al. (2017) presented listeners with implausible sentences such as "the mother gave the candle the daughter", spoken either in native English or with a strong foreign accent. Participants answered a comprehension question about each sentence according to what they thought the speaker intended to convey. With the accented speaker, listeners more often recovered the non-literal alternative "the mother gave the candle *to* the daughter" rather than the literal reading. The effect held for alternatives one edit away from the sentence heard (as with the missing "to" in the "candle" example) and was absent for active/passive alternations, which need two edits. Such accent effect was interpreted as an expectation that a less proficient speaker's utterances are a less faithful guide to what they meant.

Comprehenders track not only how much to trust an utterance but also which errors to expect from its source. Ryskin et al. (2018) asked readers to retype sentences presented as transcriptions of someone's speech and to correct any wording the speaker seemed not to intend. The surrounding sentences differed between groups of readers: for one group they contained exchanged words ("rescued by the fireman *in the time of nick*") and for another a mixture of error types. On an implausible test sentence such as "the oven cleaned the grandmother", readers who had seen exchanged words more often reversed the word order to "the grandmother cleaned the oven" than readers who had seen the mixture. This suggests that comprehenders model the kind of noise a particular source produces and not only its rate.

## 2.2 Morphosyntax

If listeners trust a linguistically less competent speaker's utterances less at the level of whole sentences, the same phenomenon should appear at the level of grammar. In studies using EEG, a morphosyntactic violation from a native speaker elicits a late positivity termed the P600, commonly interpreted as an index of reanalysis or repair. Hanulíková et al. (2012) recorded brain potentials while native Dutch listeners heard sentences spoken by a native Dutch speaker and by a Turkish speaker of Dutch. Some sentences contained a gender-agreement error, the wrong form of the definite article for the noun's gender, as in "het cultuur" for "de cultuur". Such errors are typical of non-native speakers of Dutch. The error elicited the P600 from the native speaker but not from the non-native speaker. This suggests that an error a speaker is expected to make is not processed as something to be repaired. The same study contained control sentences with semantic violations of world knowledge, such as “een dikke avond” (a thick evening) in place of “een dikke deken” (a thick blanket) in a sentence about what was put on a bed on a cold night. These violations elicited an N400, the negativity that an unexpected word elicits, and its size did not differ reliably between the speakers. Such intact N400 was interpreted as evidence against a general integration problem in accented speech.

The reduction of the late positivity also depends on which kind of errors a speaker is expected to make. In a study by Caffarra and Martin (2019), native Spanish listeners heard sentences from three native Spanish speakers and three British speakers of Spanish while their EEG was recorded, and they answered comprehension questions. Each speaker introduced himself by name and country first. The sentences contained either a gender-agreement error or a number-agreement error. In the gender error the article had the wrong gender for the noun, “la color” for “el color”. In the number error a plural article preceded a singular noun, “los color”, the equivalent of “*these* color”. English speakers of Spanish produce the first kind of error persistently and the second far less often. From the native speakers, both errors elicited the P600. From the accented speakers, the typical gender error elicited an N400 in place of the P600, and the atypical number error elicited a P600 in a later time window than in native speech. This suggests that listeners withhold the repair response selectively, for the errors their model of that class of speaker anticipates.

### 2.3 Lexical-semantic access

The speaker’s identity can bias which meaning of an ambiguous word is accessed by the listener. Cai et al. (2017) played British listeners recordings of words such as “bonnet” in either a British or an American accent and asked them to write the first word that came to mind. Such words have a different dominant meaning in the two dialects: “bonnet” is an engine cover for a British speaker and a hat for an American one. Listeners more often responded with a word related to the engine-cover meaning after a British accent and with one related to the hat meaning after an American accent. In a further experiment the critical words were morphed to be accent-neutral and were embedded among other words spoken in one accent or the other. Listeners in an American accent context still accessed the American meaning more often than listeners in a British accent context. This suggests that listeners maintain a model of the partner when interpreting their utterances rather than merely being influenced by the accentual cues in the word itself. Listeners can also hold multiple partner models at once, shifting meaning access toward the dialect of whoever is currently speaking when multiple interlocutors take turns in random order (Cai, 2022).

If a partner model can determine which sense of a word was meant, it should also determine which word a speaker was likely to choose. Wu et al. (2024) asked Mandarin listeners to judge whether a spoken word matched a picture. The words were spoken in a child's voice or in an adult's voice. A speaker first established one label for an object and then switched to a dispreferred synonym in a later phase of the session, as when a speaker who has been saying "bucket" switches to "pail". The switch elicited a larger N400 from a child voice than from an adult voice. The larger cost for the linguistically less competent speaker was attributed to an expectation that children are less flexible than adults in adopting an alternative label.

Lev-Ari (2015) had listeners follow spoken instructions to click pictures on a screen while their eye movements were tracked. Listeners first clicked a witch, a man on a magic carpet, and Santa. Imaginary creatures thus became the established theme of the display. The critical instruction was "click the *ferry*". In the dialect of the region tested, "ferry" and "fairy" sound the same; no fairy was pictured, but the display contained a mermaid. The instructions came either from a native speaker or from a non-native speaker who had made typical errors while explaining the task, such as "there have" for "there are". With the non-native speaker, listeners more often clicked the mermaid rather than the ferry, settling on the theme-consistent interpretation rather than the intended one. The anticipatory eye movements toward the theme-consistent object were graded by working memory, higher-span listeners shifting more. Such adjustment is difficult to reconcile with an account in terms of difficulty in the utterance signal alone, which predicts no role for working memory, and is better interpreted as an effortful shift toward the expectation of the listener.

### 2.4 Content expectations

The evidence in Section 2.1 was about which message a listener recovers from an implausible utterance. The evidence here is about what content a listener expects from a speaker, and about how anomalous content is processed. In an EEG study by Romero-Rivas et al. (2015), Spanish listeners heard sentences from native and from foreign-accented speakers of Spanish. The sentences included semantic anomalies such as "mi desayuno favorito es tostadas con mermelada y un *hospital* con mucha leche" (my favorite breakfast is toast with marmalade and a hospital with a lot of milk). The anomaly elicited a late positivity in the native speech but not in the accented speech. The N400 to the same anomalies was present under both accents, and it was larger and more broadly distributed in the accented speech. The absent positivity suggests that listeners did not attempt to repair an anomaly from an accented speaker, and the larger N400 is consistent with accented speech being harder to integrate.

The speaker's identity is also used at the level of sentence meaning. Van Berkum et al. (2008) recorded brain potentials from Dutch listeners hearing sentences from different speakers, with no task beyond listening for comprehension. In some sentences the content mismatched what the voice suggested about the speaker's sex, age, or social class. An example is "if only I looked like *Britney Spears* in her latest video" heard in a male voice. Such mismatches elicited an N400 beginning within a few hundred milliseconds of the critical word. This suggests that the listener's model of the speaker is consulted by the same early process that builds sentence meaning from the words. A late positivity without the early negativity has been reported for the same kind of mismatch in German (Lattner & Friederici, 2003) and, at the group level, in a Spanish adaptation of Van Berkum et al.'s design (Foucart et al., 2015).

What distinguishes the two response profiles may be the type of mismatch rather than the stage of processing. Wu and Cai (2026b) tested this contrast with native Mandarin listeners, who heard sentences spoken in voices that differed in sex and in age while their EEG was recorded. Some contents violated a social stereotype for the voice, as in "I'm going to have a *manicure* this weekend" in a male voice. Others violated biological knowledge, as in "the first time I got *pregnant* I had a hard time" in a male voice. The stereotype violations elicited an N400. The biological violations elicited a late positivity instead. The N400 was interpreted as effortful integration of unexpected content and the positivity as an attempt to correct a perceived error, in the utterance or in the perceived identity of the speaker. An anomaly may therefore be read as content to make sense of, or as an error in the utterance to correct, depending on how far the speaker is trusted.

### 2.5 Pragmatic inference

The pragmatic evidence is in some ways the clearest, because pragmatic inferences are optional in a way that lexical access is not. Hearing "pick up the *tall* cup", a listener ordinarily infers that the modifier "tall" draws a contrast with a short cup. The listener therefore fixates the tall cup sooner when a short cup is also in the display than when it is not. Grodner and Sedivy (2011) tracked listeners' eye movements as they followed such instructions from one of two speakers.

One speaker received no special introduction. The other was described as having an impairment that causes language and social problems, and during the session that speaker also mislabeled objects and produced persistently redundant descriptions. Listeners hearing the first speaker showed the usual contrastive inference. Listeners hearing the second showed no sign of it. This suggests that the inference is withheld when the modifier is not believed to have been chosen for a reason. The belief can also emerge from the speaker's behavior alone: the contrastive advantage was again no longer detectable when listeners received no description of the speaker and the speaker merely over-modified, under-modified, and mislabeled objects during the session (Gardner et al., 2021).

The most informative pragmatic evidence is from under-informativeness, as when a speaker says "*some* people have noses with two nostrils" where "all" would have been true and more informative. Fairchild and Papafragou (2018) presented native English speakers with a short biography of a speaker followed by a series of that speaker's written sentences, each to be rated for how much sense it made. The biography presented the speaker as either a native or a non-native speaker of English. Under-informative sentences were rated as making more sense when they were attributed to the non-native speaker. In a further experiment two speakers were both described as having moved from China, and only the one described as strongly accented was rated leniently, so the leniency tracked described accent strength rather than stated national origin. Accent strength was interpreted as a marker of presumed command of the language. Participants likewise ascribed a non-native speaker's omission to inability rather than unwillingness about twice as often as a native speaker's, in written justifications of why a speaker who had looked in a refrigerator said "there are bananas and apples" without mentioning the pears (Fairchild et al., 2020).

Disfluency cues are reweighted by speaker identity in the same way. In Bosker et al.'s (2014) eye-tracking experiments, Dutch listeners heard spoken instructions referring to one of two objects, one with a high-frequency name and one with a low-frequency name. The speaker was introduced as a native speaker in one experiment and as a non-native speaker in the other, and the non-native speaker imitated the native recordings sentence by sentence so that the two sets differed chiefly in accent. A native speaker's "uh" led listeners to look in anticipation toward the object whose name is harder to retrieve (e.g., a sewing machine rather than a hand). A matched instruction produced by the non-native speaker elicited no such anticipation. The difference was attributed to listeners regarding non-native disfluencies as worse predictors of the upcoming word. Listeners also showed no evidence of learning a reversed pattern, in which “uh” reliably precedes the easier word, from a non-native speaker while learning it readily from a native one (Bosker et al., 2019). The typical pattern was learned even from the non-native speaker, so the failure was specific to the reversal. The same reweighting extends to a clinical source. An editing phrase such as “the chef reached for some salt *uh I mean*” ordinarily leads listeners to anticipate the closest semantic associate of the word replaced, on the interpretation that the associate belongs to a contrast set (Lowder & Ferreira, 2016). That anticipation was attenuated when the speaker produced the sentences with a mild-to-moderate stutter (Lowder et al., 2020).

What listeners generalize from a speaker’s observed usage is nonetheless selective. Pogue et al. (2016) exposed listeners to two speakers who instructed them about displays containing a size contrast, such as a big and a small cake. One speaker said “click on the *big* cake”, whereas the other said “click on the cake” even though both cakes were present. Listeners then read new transcribed instructions and judged which speaker was more likely to have produced each one. They credited modified instructions to the appropriately modifying speaker when the adjectives were new, such as “tall” or “skinny”, with no detectable difference from the adjectives heard during exposure. In a follow-up, some new instructions carried a color adjective that failed to pick out a unique referent, such as “click on the *green* bottle” with two green bottles in view. Listeners credited these under-informative instructions to the speaker who had earlier omitted a needed adjective, so what they had generalized was an expectation about the speaker’s informativeness rather than a habit of using particular words. Exposure to a speaker who over-modified, in contrast, did not produce the same generalization about informativeness. Attribution of linguistic competence therefore appears constrained by prior expectations about how speakers ordinarily behave, rather than being a uniform adjustment.

Taken together, the comprehension evidence shows that the speaker’s identity set which meaning a form carried (e.g., Cai et al., 2017) and which message contents were expected from that speaker (e.g., Van Berkum et al., 2008). The reliance pattern involves cases where a speaker’s forms are in doubt, as with accent and non-nativeness. The less linguistically competent a comprehender believes such a speaker to be, the less they trust the utterance and the more they fall back on what they expected. The pattern appears at each level reviewed above.

Where instead the speaker's identity fixes which form they are expected to keep to, as with the child label switch, the cost of an unexpected word rises rather than falls. We explain that division in Sections 4.2 and 4.7.

## 3. Production

While a listener relies less on a speaker they believe to be unreliable, a speaker instead expends additional effort for an addressee they believe to be less able to understand them. They say more, say it more explicitly, and articulate it more clearly. This is one strand of audience design, in which speakers tailor both content and delivery to a particular addressee (Bell, 1984; Clark & Murphy, 1982). Our concern is the narrower question of how an addressee's perceived linguistic competence governs that effort.

### 3.1 Alignment

The most studied form of this additional effort is the reuse of the addressee's own words. Interlocutors converge on shared linguistic choices as a dialogue proceeds. In a maze game studied by Garrod and Anderson (1987), spoken partners settled on a common scheme for describing locations, a tendency those authors called entrainment and later work termed alignment. On one influential account, much of this convergence is a low-cost and largely automatic consequence of priming (Pickering & Garrod, 2004). That account describes modeling the partner's mind as an optional and costly strategy, invoked mainly when the automatic mechanisms fail. Alignment is, however, also partner-specific. Speakers form conceptual pacts: temporary agreements with a particular partner about how a referent is to be conceptualized (Brennan & Clark, 1996). A new expression for a familiar referent was harder to comprehend from the partner who established the original label than from one with whom no pact existed (Metzing & Brennan, 2003). Speakers also tailor referring expressions to what they believe an addressee knows (Isaacs & Clark, 1987). The studies below address this partner-specific component.

Most of the studies on linguistic competence use a naming-and-matching paradigm. A partner names a picture using one of two acceptable labels, for example "bus" or "coach", and the measure is how often the participant later uses that partner's label. Speakers align more with a partner they believe to be less able to understand than with one they believe to be more able. Cai et al. (2021) paired native Mandarin speakers with a partner in an online naming-and-matching game; the partner was in fact a set of recordings and was presented as a native adult, a native child, or a non-native adult. On critical trials the partner named a picture with either the preferred or the less preferred of its two labels, and the measure was whether the participant later used the same label for that picture. Participants aligned more with the child and with the non-native adult than with the native adult. The increased alignment was not attributable to the memorability of the recorded names or to speech rate. This suggests that speakers invest more in being understood by a partner believed to be linguistically less competent. The same increase has been found for a partner believed to be a computer rather than a human, in typed and spoken dialogue alike (Branigan et al., 2011). It has also been found for a computer presented as a basic rather than an advanced system, even though the two systems behaved identically (Branigan et al., 2011; Pearson et al., 2006). The non-native effect has also been obtained in English, both with speakers shown no evidence of their partner's performance (Ivanova et al., 2021) and with a scripted confederate in a route-giving task in which the addressee's nativeness mattered but the speaker's own did not (Suffill et al., 2021). Unscripted conversation shows the investment on a coarser measure. In a picture-card matching task, mixed pairs of native and non-native speakers used nearly twice as many words as native pairs, and the expressions native speakers produced for non-native partners were rated as less idiomatic by native judges (Bortfeld & Brennan, 1997). Convergence on shared expressions in those pairs did not detectably differ from that in native pairs.

The same pattern appears at the level of how a referent is conceptualized. In two experiments by Y. Zhang and Cai (2026), Mandarin speakers took turns with a partner describing pictures of people so that the partner could click on the picture described. The partner was a child or an adult in one experiment and a non-native or a native speaker in the other. The partner consistently referred to a doctor as "the white coat" rather than "the doctor", describing people by their appearance rather than their occupation. Speakers adopted that perspective on new pictures of their own, describing a lawyer as "wearing a tie" rather than as "a lawyer". They did so more for the child than for the adult partner and more for the non-native than for the native partner. The adjustment was larger in this communicative version than in a non-communicative version of the task, in which a staged technical failure meant the partner could not hear them and no matching was required. These effects were interpreted as an implicit adjustment together with a strategic component that an explicit goal amplifies.

One case runs in the opposite direction: in typed dialogue, speakers who were themselves language learners aligned more with a native than with a non-native partner (Shen & Wang, 2025; D. Zhang & Nicol, 2022). The pattern appears to track the partner's native status rather than relative proficiency, since alignment showed no detectable difference between non-native partners described as more and as less proficient than the speaker (Shen & Wang, 2025).

If greater alignment followed from a partner's being a machine as such, it should extend to large language models. Recent evidence suggests it does not. In typed conversation, users converged lexically and stylistically toward the language models of one provider at rates broadly consistent with human-to-human text corpora (Blevins, 2026). In spoken dialogue, one study found reduced rather than increased structural alignment when the same confederate was believed to be artificial (Li, 2025). The beliefs that produced the earlier computer-alignment effects were beliefs about limited, unsophisticated systems. If the behavior tracks the belief, it should change as the belief changes. Contemporary language models are not generally believed to be poor language users, and the earlier pattern should weaken accordingly.

### 3.2 Referential choice

Alignment involves which of the partner's words a speaker reuses. A second and independent adjustment involves how much a speaker says. Tal et al. (2023) recorded parents narrating a picture book to their own children aged one to six. They counted how often a character was referred to with a full noun phrase or a name rather than a pronoun, the equivalent of "the boy" over "he". The younger the child, the more of these explicit expressions the parents used. In a second study reported alongside it, adult speakers told the same story to a second-language learner or to a native-speaking adult, and they used more explicit expressions for the learner. The parallel between children and adult learners suggests that the operative variable is perceived linguistic competence rather than age. A converging effect appears on a global measure of redundancy: speech to young children grew less repetitive over the first three years (Tal et al., 2024).

A similar increase in explicitness has been reported for machine partners, which can be believed to be less able simply because of how they are presented. People apply social expectations to computers even while knowing they are not human (Nass & Moon, 2000). Dunn and Cai (2025) had participants in three experiments type descriptions in a matching-and-naming task for a partner framed as a computer or as a human, and the partner either always or never included a redundant modifier in its own descriptions. In one of the three experiments participants produced more redundant descriptions overall toward the computer partner, writing “the *blue* square” when only one square was on the screen. In all three experiments participants also aligned with the partner’s usage, producing more redundant descriptions with a redundant partner than with a non-redundant one, and this alignment showed no detectable difference between computer and human partners. How much detail to add to a description thus varied with the partner’s framing in one experiment, whereas how much to reuse the partner’s own level of redundancy did not in any of the three. That null does not contradict the stronger alignment toward a computer reported by Branigan et al. (2011), since that effect involved the choice of name rather than redundancy. Dunn and Cai attributed the difference to the level of representation, regarding redundancy alignment as largely automatic and lexical alignment as partly goal-directed.

Speakers also tailor lexical choices to their partner. The relevant studies ask whether the operative representation is a model of the partner or a response to the utterance signal itself. In the paradigm of Cai et al. (2025), a partner relays the definition of a word and the speaker supplies the word. The partners differed in dialect, and the speakers were bidialectal. Speakers produced more of the variants favored in the partner’s dialect, “apartment” rather than “flat” for a partner who sounded American. The effect was no weaker on trials where the definition arrived in writing and so carried no accent information. When two partners of different dialects alternated, speakers produced the variant matching whichever partner was currently listening. This suggests that speakers maintain a top-down model of each partner rather than responding bottom-up to accent, the production-side counterpart of the comprehension finding (Cai et al., 2017).

**3.3 Acoustic register**

Explicitness is not only a matter of which words are chosen but also of how they are articulated. Speakers tune the acoustic form of what they say, the core of hyper- and hypo-articulation theory (Lindblom, 1990). On that theory, speech varies from hypospeech to hyperspeech according to the speaker’s estimate of how much the listener can supply from sources independent of the utterance signal. Without demands on the utterance, articulation defaults to a low-cost form; the criterion is discriminability sufficient for lexical access. Hyperarticulation supplies the clearest evidence of adjustment to a less able listener: the vowels are pushed further apart so that the acoustic space they enclose expands. The point vowels /i/, /a/, and /u/ mark the corners of that space. English examples are the vowels of “bead”, “pot”, and “boot”. Kuhl et al. (1997) recorded American, Russian, and Swedish mothers speaking to their two- to five-month-old infants and to an adult native speaker. The target words carried these three vowels. Every one of the thirty mothers produced the three vowels further apart when speaking to her infant than when speaking to the adult. This is the adjustment hyper- and hypo-articulation theory predicts for a listener who can supply little from sources other than the utterance. Hyperarticulation has also been reported in both infant-directed and foreigner-directed speech relative to ordinary adult-directed speech, with no detectable difference between the two (Uther et al., 2007). The raised pitch characteristic of infant-directed speech, however, was not extended to foreigners. Infant-directed and pet-directed speech both raised pitch and affect relative to adult-directed speech, but only the former hyperarticulated (Burnham et al., 2002). The separation between affect and articulation suggests that the articulatory adjustment tracks the addressee’s linguistic needs rather than merely an affective stance.

Speech to voice interfaces is a clear test case: a device can be judged to have low linguistic competence on the basis of belief alone. In recordings of children and adults instructing a voice assistant and a human experimenter, Cohn, Barreda, et al. (2024) found a distinct device-directed register in utterance duration and pitch. Children raised pitch more for the device and adults lengthened utterances more; after a staged recognition failure, speakers raised pitch further, most clearly the children. The children’s adjustment was interpreted as reflecting their conception of the system’s ability to understand rather than as an anthropomorphism response.

Taken together, the production evidence shows the complementary pattern: the less linguistically competent a speaker believes an addressee to be, the more they invest in the utterance. The investment shows in more explicit reference, more alignment with the partner’s own words, and clearer articulation. The production manipulations act mainly on the decoding side: a foreign accent, a child’s age, or a device’s presentation each lowers how well the addressee is expected to recover a message. The comprehension manipulations divide differently. For non-native and accented speakers, the cue is the reliability of their forms; for artificial partners it is mainly their knowledge; and for children it is both the forms and contents expected of them and the reliability of their forms. In Section 4.6 we state the assumption that one fidelity serves both modes, and in Section 4.7 we locate the three partner types in a two-quantity space of fidelity and knowledge.

## 4. A model of rational interlocutors

The evidence reviewed in Sections 2 and 3 leaves the discounting-vs-investing paradox in place: one belief about a partner makes comprehenders rely less on the utterance and producers invest more in it. In this section we develop the rational interlocutor (RI) model, which derives both adjustments from a single computation. It specifies what a rational interlocutor should compute given a model of their partner. We build the computation in comprehension first: the choice rule itself (Section 4.1), the channel and the partner's fidelity (Section 4.2), and the message prior and the partner's knowledge (Section 4.3). In Section 4.4 we then state the resulting partner model as a whole and derive what each of its three beliefs does. Production is the same computation run in the opposite direction (Section 4.5); in Section 4.6 we set the two modes side by side, and in Section 4.7 we apply the model to three partner types.

Because listener and speaker change their reference between comprehension and production, we use two other names throughout the paper. We call the person whose processing (comprehension or production) is being described by the RI model the *agent*. We call the person described by the three parameters ($\Pi$, $\Phi$, and $\Lambda$) the *partner*. In comprehension the partner is the one speaking and the agent the one listening; in production the agent is the one speaking and the partner the one listening. Where the direction matters, we write *speaking partner* and *listening partner*. Table 1 lists the notation.

**Table 1.** *Notation.*

| Symbol | Meaning |
|---|---|
| $m, u$ | a *message* and the *utterance* that carries it |
| $\Pi$ | *partner's identity*: the agent's beliefs about the partner's age, gender, social background, dialect background, region of origin, familiarity, relationship, interactional role, and whatever else affects what they are expected to mean, the forms they use, and what is appropriate to say to them |
| $\Phi$ | *partner's fidelity*: the agent's belief about how likely the partner's utterances are to keep to the canonical forms of their messages, both when the partner produces an utterance and when they recover a message from one |

| Symbol | Meaning |
|---|---|
| $\Lambda$ | *partner's knowledge*: the agent's belief about how likely the partner's messages are to keep to what a knowledgeable speaker would mean, given their knowledge of the world |
| $P(m;\Pi,\Lambda)$ | *message prior*: how likely this partner is to mean message $m$. $\Pi$ sets the typical messages, and $\Lambda$ sets how likely a message is to keep to them |
| $P(u \mid m;\Pi,\Phi)$ | *channel*: how a message is realized as an utterance for this partner. $\Pi$ sets the canonical form of the message, and $\Phi$ sets how likely the utterance is to keep to it |
| $c_{\Pi}(m)$ | *canonical form*: the form a partner of identity $\Pi$ typically uses for message $m$; the center of the channel's faithful component |
| $P(m \mid u;\Pi,\Phi,\Lambda)$ | *recovery*: the probability that the partner arrives at message $m$ on hearing utterance $u$; by Bayes' rule it is the channel weighted by the message prior and normalized over the candidate messages |
| $P(u;\Pi)$ | *utterance prior*: how available or appropriate an utterance is, combining the agent's own standing preference for fluent forms with the typical-utterance distribution of a partner of this identity |
| $s(x)=-\log P(x)$ | *surprisal*: how unexpected a message or an utterance is |
| $m^{*},u^{*}$ | the *chosen* message in comprehension and utterance in production |
| $\alpha$ | *rationality parameter*: how sharply the best option is preferred |
| $\sigma$ | *width of a Gaussian component* |
| $P_{\text{typ}}(m;\Pi)$ | *typical-message distribution*: the typical messages of a partner of identity $\Pi$, what a knowledgeable partner of that identity would be expected to mean |
| $P_{\text{typ}}(u;\Pi)$ | *typical-utterance distribution*: the forms in use with a partner of that identity |
| $P_0(u)$ | *standing preference*: the agent's own preference for fluent, available forms, whoever the partner is |
| $k$ | *number of descriptors* in an utterance |
| $\kappa$ | *per-descriptor cost*: the effort charged for each descriptor added to an utterance, fixed as 0.05 in the illustrations that use it |
| $\omega$ | *partner weight*: the degree to which the agent takes the partner into account |

*Note.* The semicolon separates a distribution's variables from the parameters that fix it. $\Pi$, $\Phi$, and $\Lambda$ are held at point values in the account as stated rather than as distributions the agent maintains over them. We state the conventions governing the symbols in Section S1 of the supplementary materials.

**4.1 Comprehension as inference over a noisy channel**

The utterance that reaches a comprehending agent is an imperfect guide to what the partner meant. Partners misspeak, drop words, and substitute one word for another. A rational agent therefore does not take the utterance at face value, but asks which intended message most likely gave rise to it. In “the mother gave the candle the daughter”, comprehenders often recover the non-literal alternative “the mother gave the candle *to* the daughter”, as if a small word “to” had gone missing (Cai et al., 2022; Gibson et al., 2013). The utterance is read not as the message itself but as evidence about it.

The noisy-channel model formalizes this inference. Levy (2008) states it over word strings under uncertain perceptual input, and Gibson et al. (2013) state it over sentence strings, reading meanings off them; we state it over messages. We write $m$ for a candidate intended message and $u$ for the utterance received. The agent wants the message most probable given the utterance, $P(m \mid u)$, and Bayes’ rule makes it proportional to the product of two other quantities:

$$P(m \mid u) \propto P(u \mid m) P(m). \tag{1}$$

The first quantity $P(u \mid m)$ is the *channel*, the noise model of Gibson et al. (2013): how likely an intended message is to surface as the utterance received, faithfully or garbled. The second, $P(m)$, is the *message prior*: how likely that message was to be meant before the utterance is heard. In the “candle” example, the non-literal reading has a high message prior, since transfer between two people is a common event. The channel favors the literal reading, which needs no edit, but only slightly because dropping one short word is a common error. The message prior therefore decides in favor of the non-literal reading.

In our model we let $m$ and $u$ stand for messages and utterances in general rather than only whole sentences: a single word, a referring expression, or a sound can play the role of $u$. We then make both quantities depend on the particular partner through three parameters. The channel depends on the partner’s identity $\Pi$ and fidelity $\Phi$: the identity sets the *canonical form* $c_{\Pi}(m)$ of each message, and the fidelity sets how reliably their utterances keep to it (Section 4.2). The message prior depends on the same identity $\Pi$ and on the partner’s knowledge $\Lambda$: the identity sets the typical messages, and the knowledge sets how likely a message is to keep to them (Section 4.3). Together the three define the *partner model*: the agent’s beliefs about this partner that are relevant to communicating with them (Section 4.4). Parameterizing (1) accordingly gives

$$P(m \mid u; \Pi, \Phi, \Lambda) \propto \underbrace{P(u \mid m; \Pi, \Phi)}_{\text{channel}} \underbrace{P(m; \Pi, \Lambda)}_{\text{message prior}}. \tag{2}$$

From here on, *channel* and *message prior* refer to these partner-dependent quantities: $P(u \mid m; \Pi, \Phi)$ is how this particular partner turns a message into an utterance, and $P(m; \Pi, \Lambda)$ is the agent's expectation of what this particular partner will mean. Both are usefully restated in terms of *surprisal*, which converts a probability into a cost. The surprisal of an outcome of probability $p$ is

$$s = -\log p, \tag{3}$$

which is zero when the outcome is certain and grows without bound as it becomes rare. In plain terms, surprisal is the degree to which an outcome departs from expectation: an outcome already taken for granted costs nothing to accept, whereas one that was barely credible carries a large cost. Taking logarithms also turns the product in (2) into a sum. We write $s(u \mid m; \Pi, \Phi) = -\log P(u \mid m; \Pi, \Phi)$ for how surprising the received utterance would be if the partner had meant *m*, and $s(m; \Pi, \Lambda) = -\log P(m; \Pi, \Lambda)$ for how surprising the message *m* is for this partner to mean in the first place.

An agent settling on the single most probable message selects the *m* that makes the total of these two costs as small as possible:

$$m^* = \arg\min_m [\underbrace{s(u \mid m; \Pi, \Phi)}_{\text{mismatch cost}} + \underbrace{s(m; \Pi, \Lambda)}_{\text{message cost}}]. \tag{4}$$

Here $\arg\min_m$ means the message that makes the bracketed total smallest. We term $s(u \mid m; \Pi, \Phi)$ the *mismatch cost*, which is high when the received utterance would be an unlikely way for the partner to express the candidate message. It grows with the distance between the utterance received and the canonical form of the candidate message for this partner. We term $s(m; \Pi, \Lambda)$ the *message cost*, which is high when the message is unlikely for this partner to have meant. Because making a sum of surprisals small is the same as making the corresponding probabilities large, (4) can equally be written as choosing the most probable message:

$$m^* = \arg\max_m [\log P(u \mid m; \Pi, \Phi) + \log P(m; \Pi, \Lambda)], \tag{5}$$

the form we carry to production. On either formulation, comprehension is not a passive reading-off of the utterances: the interpretation that best matches them is resisted when it would be implausible for this partner to mean.

**4.2 Fidelity and the channel**

The balance between the two costs depends on the shape of the channel, and two beliefs about the partner set that shape. The channel $P(u \mid m; \Pi, \Phi)$ is a belief about *which form* a message takes for this partner, and about *how reliably* the utterance keeps to that form. The identity $\Pi$ sets the form: for each message $m$ it fixes the *canonical form* $c_\Pi(m)$, the form a partner of this identity typically uses for it (Equation 6). The fidelity $\Phi$ sets how likely the utterance is to keep to the canonical form. We identify $\Phi$ with *the sharpness of the channel*, the agent's belief about how reliably forms and meanings map onto each other for this partner. It covers the lexicon and the grammar with its morphology as well as the articulation. What belongs to $\Phi$ is the reliability in the partner's use of any of these, not a consistent difference from the agent's own usage; which mapping it is belongs to $\Pi$ (Section 4.4, Figure 2).

Where the agent believes a partner's fidelity to be high, each message maps to just a few likely utterances and each utterance back to just a few likely messages, so the mismatch cost discriminates between candidates. Where fidelity is low, many utterances are compatible with a message and many messages with an utterance. The mismatch cost then barely changes from one candidate to another, and the message that minimizes the total in (4) would be the one the prior already favored. Discounting a linguistically less competent partner's utterances therefore follows from flattening the channel term.

Equations 4 and 5 fix how the two costs combine but leave the channel itself unspecified. We fix a minimal definite form here and use it for every illustration. Messages and utterances are points on a shared one-dimensional scale: nearby points stand for messages easily confused, and distant points for messages that are not. The channel is a mixture of a faithful and an unfaithful component,

$$P(u \mid m; \Pi, \Phi) = \underbrace{\Phi\, N(u; c_\Pi(m), \sigma^2)}_{\text{faithful}} + \underbrace{(1 - \Phi)\, \text{Uniform}(u)}_{\text{unfaithful}}, \tag{6}$$

so that with probability $\Phi$ the message comes out close to its canonical form $c_\Pi(m)$, and otherwise comes out as a form unrelated to the message, every form on the scale being equally likely. The computed illustrations (Figures 1 and 3 to 5) set $c_\Pi(m) = m$ for every partner, the message's own point on the shared scale. With $\Phi$ at 0.9, for instance, nine utterances in ten come out close to the canonical form of the message and one in ten comes out as an unrelated form. $\Phi$ is the weight the agent assigns to the partner encoding and decoding as a fully competent interlocutor would. The space is bounded and each component is a proper density on it, the Gaussian being truncated and renormalized before it is mixed, so the mixture is itself proper and $\Phi$ is exactly the weight on the faithful component. $\Phi$ is not interchangeable with the width $\sigma$ of that component, and the difference shows in how discrimination depends on distance. Lowering $\Phi$ mixes in a floor that no separation between messages overcomes: however far apart two candidates lie, the utterance can favor one over the other only up to a bound set by $\Phi$. Widening $\sigma$ blurs nearby candidates but leaves discrimination growing without bound as candidates move apart. A partner may sometimes drop a word or substitute an unrelated one; that flattens the faithful component (lowers $\Phi$): even “ship” and “elephant” are then only partly distinguishable, because either could be a random substitution for the other. A partner’s accent may instead make “ship” and “sheep” hard to tell apart; that widens the component (increases $\sigma$): the two forms that differ in one vowel blur together, whereas “ship” and “elephant” remain fully distinguishable. The account as stated holds $\sigma$ fixed and varies only $\Phi$. A consistent accent displaces or broadens that component rather than flattening it (Kleinschmidt & Jaeger, 2015). Equation 6 represents the displacement through $c_{\Pi}(m)$ and the broadening through $\sigma$. Dialect-specific forms are a displacement of this kind: for a British partner the canonical form of the engine-cover message is “bonnet” and for an American one “hood”, so the identity moves the faithful component while the fidelity is unchanged. Accent and fidelity are therefore separable in principle even where they are often confounded in practice. The “bonnet” case is the identity acting in the channel, on the canonical form. The identity acts in the message prior as well, in the “foundation” case of Section 4.3, and in Section 4.4 we set the two effects side by side.

Equation 6 is an *implementation equation*: it fixes a definite form and definite values for illustrations, as a choice of convenience rather than a claim of the account. The other implementation equations are Equation 7; Equations 13 and 14; Equations 16 and 17; Equations 18 and 19; and Equation 21. Every illustration is computed from these alone. Table S1 of the supplementary materials lists all the equations of the paper and gives the values at which each implementation equation is run. As is shown in Figure 1, the utterance favors $m_1$ over $m_2$ by odds of about seventy to one in the sharp case and about four to one in the flat one. In the “candle” example, $m_1$ is the literal reading and $m_2$ the non-literal reading. The canonical form of the first is the utterance heard, and that of the second is one word away. A sharp channel makes the utterance strong evidence for the literal reading; a flat one makes it weak evidence, and the message prior then decides.

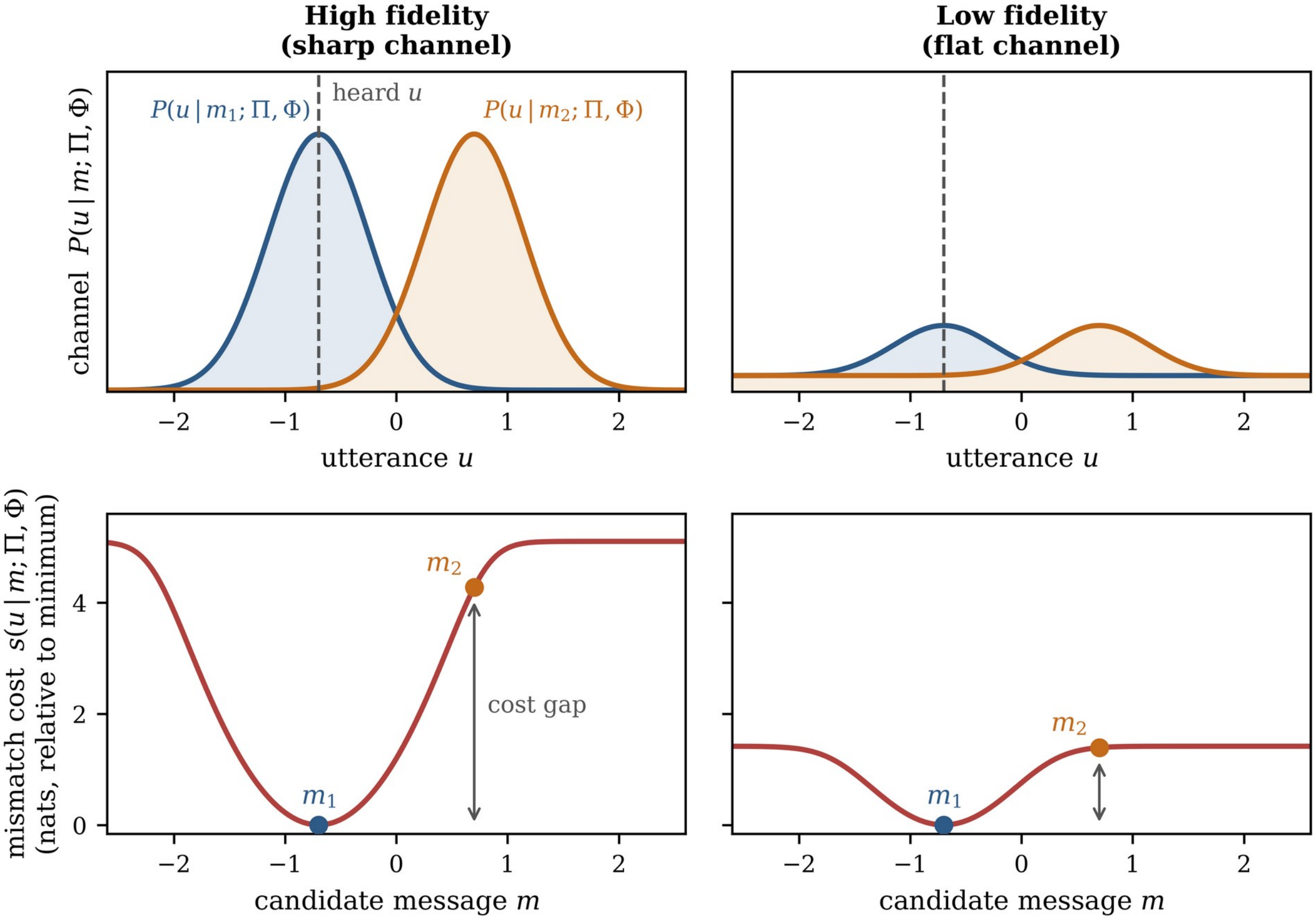

**Figure 1.** *Fidelity as the sharpness of the channel.* The horizontal axis is the shared message-utterance scale of Equation 6, utterances in the top row and candidate messages in the bottom row; nearby points are easily confused. Top row: the channel of Equation 6 for two intended messages $m_1$ (blue) and $m_2$ (orange); high fidelity on the left, low on the right; the flat floor at low fidelity is the unfaithful component. Bottom row: the mismatch cost $s(u \mid m; \Pi, \Phi)$ of each candidate message given the utterance at the dashed line, relative to its minimum; the arrow marks the cost gap, 4.27 nats (left) and 1.39 nats (right). $\Phi = 0.92$ (left); $\Phi = 0.18$ (right); $\sigma = 0.45$; $m_1 = -0.7$; $m_2 = 0.7$; the heard utterance $u = -0.7$; the space is $[-8, 8]$, plotted from -2.6 to 2.6.

The noisy channel has already been used to explain how a linguistically less competent language user comprehends, so locating a partner's encoding fidelity in its sharpness is not a completely new mechanism. In a commentary, Futrell and Gibson (2017) attributed non-native comprehension differences to a higher assumed rate of error in the input and to a less precise probabilistic model of the grammar. The assumed error rate flattens the channel, and it is a belief about channel fidelity of precisely the kind specified here. The less precise grammar flattens their prior over sentence forms instead. We place the grammar in the mapping from messages to forms, so that factor too belongs to the channel. They analyze the non-native person's *own* comprehension, so the elevated noise is located in the comprehender, whereas Gibson et al. (2017) place it in a belief about an accented partner. The same logic has been applied to comprehension in aphasia, again on the side of the impaired comprehender. People with aphasia are argued to comprehend with the same rational inference as everyone else but to assume more noise, relying less on the literal syntax of the input and more on priors favoring frequent constructions and plausible events (Gibson et al., 2016).

**4.3 Knowledge and the message prior**

The message prior $P(m;\Pi,\Lambda)$ mirrors the channel. The channel says which utterance to expect for a message; the message prior says which message to expect from this partner in the first place. The same two kinds of belief set its shape (Section 4.4, Figure 2). The identity sets the typical messages, and the knowledge sets how likely the partner's message is to keep to them, as the fidelity sets how likely an utterance is to keep to the canonical form.

On a probabilistic reading, $\Lambda$ is the probability the agent assigns to the partner's message being drawn from the constrained set of things a knowledgeable speaker would mean. Both $\Phi$ and $\Lambda$ then behave as mixture weights. $\Phi$ is the weight on the channel's faithful component (Equation 6). $\Lambda$ is the weight on the constrained component of the message prior, the component that $P_{\text{typ}}(m;\Pi)$ supplies,

$$P(m;\Pi,\Lambda)=\underbrace{\Lambda\,P_{\text{typ}}(m;\Pi)}_{\text{constrained}}+\underbrace{(1-\Lambda)\,\text{Uniform}(m)}_{\text{unconstrained}}, \tag{7}$$

so that with probability $\Lambda$ the partner means one of the things a knowledgeable speaker would mean, and otherwise means something unconstrained by that knowledge, any message being as likely as any other. Equation 7 does for the message prior what Equation 6 does for the channel, and the illustrations compute every message prior from it. Within the message prior, $\Pi$ sets what is typical and $\Lambda$ how likely the partner is to keep to it. $\Pi$ sets the typical message: for example, the word "foundation" has the same form for every partner, but a female voice makes the cosmetic the more typical message and a male voice the base of a building. The typical-message distribution therefore differs by identity while the channel does not. $\Lambda$ sets how likely the partner is to keep to the typical message: for example, a message about drug interactions is typical for a physician. For an experienced physician the agent holds the messages tightly to what a knowledgeable physician would mean, so a medically nonsensical remark from them comes as a large surprise. For a first-year medical student the identity points to the same medical topics, but the agent holds the messages to that distribution more loosely, so the same nonsensical remark surprises the agent less. As with the channel, each component is a proper density on the bounded space, so Equation 7 is itself proper and $\Lambda$ is exactly the weight on the constrained component.

The knowledge is a belief about what the partner knows about the world, and it is distinct from the fidelity. The fidelity is the reliability of the mapping between forms and meanings, not its particular forms. A partner's typical word for a given meaning belongs to $\Pi$ (Section 4.2), and $\Phi$ carries how reliably that word's form and meaning map onto each other. Knowledge in this sense is how knowledgeable the partner is believed to be, not what the two interlocutors both know and know that they know; the latter is common ground, a different construct that we discuss in Section 6.1. Neither $\Pi$ nor $\Lambda$ requires a detailed representation of the partner: the identity is in most of the studies reviewed a category-level belief (Section 5.1), and the knowledge is a single weight.

**4.4 The partner model: identity, fidelity, and knowledge**

In Sections 4.1 to 4.3 we have specified the partner model in pieces. Stated as a whole, it is defined by three beliefs held by the agent. The first is the partner's *identity* ($\Pi$): attributes such as age, gender, and dialect that set what is typical of them. These attributes set which messages the partner typically means (their *typical messages*) and which form they typically use for each message (its *canonical form*). They also set which forms are appropriate to use with them. This is the belief the speaker-modeling and audience-design literatures have long worked with. The second is the partner's *fidelity* ($\Phi$): how likely the utterance that carries a message is to keep to the canonical form of that message, in either comprehension or production. The third is the partner's *knowledge* ($\Lambda$): how likely the partner's intended message is to keep to their typical messages, those a knowledgeable speaker of that identity would mean. As shown in Figure 2, the identity sets the typical value in the channel and in the message prior; the fidelity and the knowledge set how likely the utterance and the message are to keep to that value rather than deviate from it. One computation reads these beliefs in two modes. Comprehension is the mode built so far, and production reads the same three beliefs in the opposite direction (Section 4.5).

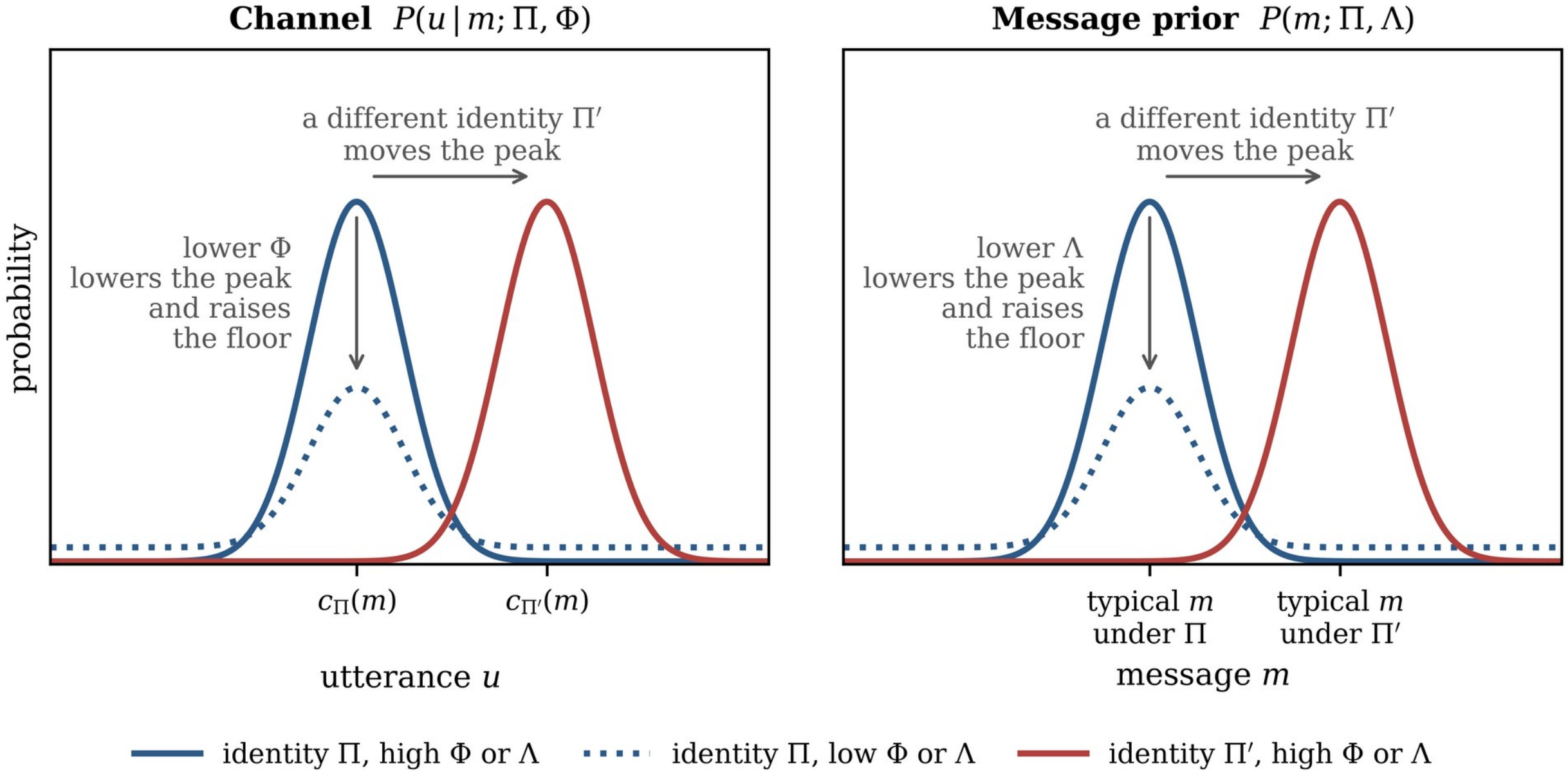

**Figure 2.** *The partner model: identity sets the typical value, fidelity and knowledge set how likely the partner is to keep to it.* Left panel: the channel $P(u \mid m; \Pi, \Phi)$ of Equation 6 over utterances for one message $m$, with the faithful component centered on the canonical form $c_\Pi(m)$. Right panel: the message prior $P(m; \Pi, \Lambda)$ of Equation 7 over messages, with the constrained component centered on the typical message. In the left panel the floor is the unfaithful component and in the right panel the unconstrained component. Blue: identity $\Pi$; red: a different identity $\Pi'$; solid: fidelity or knowledge at 0.9; dotted: at 0.4; $\sigma = 0.45$; the typical value at $-0.5$ under $\Pi$ and 1.3 under $\Pi'$; the space $[-8, 8]$, plotted from $-3.4$ to 3.4.

For two candidate messages $m_1$ and $m_2$, the logarithm of Equation 2 decomposes the log odds between them into two terms,

$$\log \frac{P(m_1 \mid u; \Pi, \Phi, \Lambda)}{P(m_2 \mid u; \Pi, \Phi, \Lambda)} = \underbrace{\log \frac{P(u \mid m_1; \Pi, \Phi)}{P(u \mid m_2; \Pi, \Phi)}}_{\text{channel log-likelihood ratio}} + \underbrace{\log \frac{P(m_1; \Pi, \Lambda)}{P(m_2; \Pi, \Lambda)}}_{\text{prior log-odds gap}}. \tag{8}$$

The first term is the *channel log-likelihood ratio*, which depends on $\Pi$ and $\Phi$. The second is the *prior log-odds gap*, which depends on $\Pi$ and $\Lambda$. The two are additive; $\Phi$ appears only in the first term and $\Lambda$ only in the second, while $\Pi$ reaches both. Each has a distinct role in the comparison. $\Phi$ scales how far the utterance received discriminates between the candidates: at $\Phi = 1$ the channel is as sharp as its width $\sigma$ allows and the utterance discriminates most, and at $\Phi = 0$ it is flat and says nothing. $\Pi$ sets which form each candidate takes for this partner, and with it which candidate the utterance received favors. The mixture of Equation 7 expands the prior log-odds gap,

$$\log \frac{P(m_1; \Pi, \Lambda)}{P(m_2; \Pi, \Lambda)} = \log \frac{\Lambda P_{\text{typ}}(m_1; \Pi) + (1 - \Lambda)\,\text{Uniform}(m_1)}{\Lambda P_{\text{typ}}(m_2; \Pi) + (1 - \Lambda)\,\text{Uniform}(m_2)}, \tag{9}$$

where the two uniform terms are equal on the bounded space. In plain terms, Equation 9 is how far the message prior favors one candidate message over the other. At $\Lambda = 1$ the gap is the log odds under $P_{\text{typ}}$ alone, so the partner is held to what a knowledgeable speaker would mean; at $\Lambda = 0$ the two mixtures reduce to the uniform terms and the gap is zero. The partner's identity $\Pi$ sets which candidate the message prior favors, because $P_{\text{typ}}(\cdot; \Pi)$ fixes which candidate is the more typical. The knowledge $\Lambda$ sets how far the gap opens toward $P_{\text{typ}}$'s log odds. The two identity effects of Sections 4.2 and 4.3 show the two terms at work. The "foundation" effect engages the second term alone. Both the cosmetic meaning and the building meaning share the same form for every partner, so the channel log-likelihood ratio is zero at any fidelity and the prior log-odds gap decides. A female voice makes the cosmetic meaning the more typical message, and the partner's believed knowledge sets how far that shifts the reading. The "bonnet" effect lies in the first term instead. The engine-cover message takes the form "bonnet" for a British partner and "hood" for an American one, so the utterance "bonnet" leaves the engine cover available for the first partner and points to the hat for the second. The two effects should therefore depend on fidelity differently. The "foundation" effect should not depend on the partner's fidelity, whereas the "bonnet" effect should weaken as fidelity falls, since a flattened channel says less about which form the partner would have used. The "candle" effect of Section 4.1 runs the other way and should grow as fidelity falls, since there the channel favors the literal reading until it is flattened.

There are three consequences following from that decomposition. First, *fidelity and knowledge compete*. The channel log-likelihood ratio grows as fidelity rises, and the prior log-odds gap widens as the partner is believed to know more. The larger of the two in magnitude decides the interpretation. The more a partner is believed to know, the higher the fidelity at which the prior log-odds gap still dominates. A partner believed to know nothing is the limiting case: their message prior is flat, so there is no expected message to fall back on, and discounting their utterance can at most leave the agent undecided between readings. Second, *knowledge gates identity in the prior*: at $\Lambda = 0$ the prior log-odds gap is zero whatever $\Pi$ is. Interpretation should then be unaffected by a manipulation of the partner's identity that acts through the message prior, as in the "foundation" case. The effect of such a manipulation should grow with $\Lambda$. The gating reaches the message prior only. An identity effect survives at $\Lambda=0$ when it is carried by the channel, as in the “bonnet” case. Third, the two terms enter the sum without interacting. In the log odds between two candidate readings in comprehension, a fidelity manipulation and a knowledge manipulation should therefore be additive.

**4.5 Production as the mirror image of comprehension**

Production is the same computation run in the opposite direction of comprehension. The agent now holds the message *m* fixed and chooses the utterance *u*, asking which utterance will best help the listening partner recover the message. The two costs are again surprisals. The first, $s(m \mid u;\Pi,\Phi,\Lambda)$, is the partner’s surprise at arriving at *m* after hearing a candidate *u*: how likely they are to fail to recover it. The second, $s(u;\Pi)$, is how unusual or effortful that utterance is for this partner: low for an utterance common and appropriate for them and high for a rare or elaborate one. The agent selects the utterance that makes the total smallest:

$$u^{*}=\arg\min_{u}[\underbrace{s(m \mid u;\Pi,\Phi,\Lambda)}_{\text{misrecovery cost}}+\underbrace{s(u;\Pi)}_{\text{production cost}}], \tag{10}$$

equivalently, choosing the utterance that makes the partner’s recovery most likely once effort is allowed for:

$$u^{*}=\arg\max_{u}[\log P(m \mid u;\Pi,\Phi,\Lambda)+\log P(u;\Pi)]. \tag{11}$$

Equations 10 and 11 mirror (4) and (5) in structure. We term $s(m \mid u;\Pi,\Phi,\Lambda)$ the *misrecovery cost*, which is the channel inverted: the surprisal of this partner’s recovering the intended message from the candidate utterance, weighed against the other messages they might reach instead. Making it small means choosing the utterance most likely to be understood, which is the locus of audience design. The recovery distribution is derived from the channel and the message prior of Equation 2 by Bayes’ rule,

$$P(m \mid u;\Pi,\Phi,\Lambda)=\frac{P(u \mid m;\Pi,\Phi)P(m;\Pi,\Lambda)}{\sum_{m'}P(u \mid m';\Pi,\Phi)P(m';\Pi,\Lambda)}, \tag{12}$$

where the sum runs over all the candidate messages the partner might mean. Equation 12 also explains an asymmetry between the two modes. In comprehension the channel term $P(u \mid m;\Pi,\Phi)$ is the forward channel and carries no $\Lambda$. In production the agent needs the partner’s decoding, which reads the channel backward. Reading it backward requires Bayes’ rule, which brings the partner’s message prior and the normalizing sum into the term, so the recovery term carries not only $\Pi$ and $\Phi$ but also $\Lambda$. We term $s(u;\Pi)$ the *production cost*, which penalizes utterances rare or inappropriate for this partner. It carries two things at once. The first is the agent's own standing preference for fluent and available forms, which is what the cost term in the rational speech act (RSA) model encodes (Goodman & Frank, 2016). A form the agent seldom uses carries a high cost of the first kind, however appropriate it would be for the partner. The second is a belief about which forms are appropriate for this particular partner, which we write as the *typical-utterance distribution* $P_{\text{typ}}(u;\Pi)$: the forms in use with a partner of that identity. A speaker can say "good afternoon" or "hi" with equal ease, so neither is harder to produce. The first is the form in use with a senior colleague and the second the form in use among friends, a difference of convention rather than of what a particular partner knows. What makes "hi" costly is who is listening, not the agent's own fluency. Likewise, "apartment" rather than "flat" is the appropriate choice for a partner who sounds American (Cai et al., 2025). Only that second part depends on the partner, and it mirrors the typical-message distribution $P_{\text{typ}}(m;\Pi)$ of the message prior. Writing the agent's own preference as $P_0(u)$, the utterance prior is $P(u;\Pi) \propto P_0(u)P_{\text{typ}}(u;\Pi)$. We hold $P_0(u)$ fixed as a background cost.

$\Pi$ and $\Lambda$ reach production through the recovery term $s(m \mid u;\Pi,\Phi,\Lambda)$, where they set what the partner will understand an utterance to mean: the probability of being understood. $\Pi$ alone also reaches it through the utterance prior $P(u;\Pi)$, where it sets what is appropriate to say to them. The knowledge $\Lambda$ does not enter the utterance prior, because which forms are appropriate for a partner depends on their identity rather than on how likely their messages are to keep to the typical messages. Whether a particular partner would recover a rare word is charged in the recovery term instead, so the utterance prior carries only the convention.

The notation may invite a misreading here. If the utterance prior were the distribution jointly implied by the channel and the message prior, $\sum_m P(u \mid m;\Pi,\Phi)P(m;\Pi,\Lambda)$, the two terms in (11) would collapse into one by Bayes' rule. The agent would then simply produce the form that best fits $m$ under the partner's own mapping from messages to forms. The design for recoverability would drop out. In the RSA model the cost of an utterance is a property of the form alone, such as its length or frequency (Goodman & Frank, 2016). We adopt that independence: the utterance prior is specified without reference to the channel, so the two terms in (11) cannot collapse. $P_{\text{typ}}(u;\Pi)$ is likewise a property of the forms, the dialect and register in use with a partner of this identity, and not a marginal of the channel over the messages they might mean.

We use the number of descriptors an agent uses as an illustration. An utterance may consist of one descriptor or of several: “the cup” has one descriptor; “the red cup on the table” has three. Choosing an utterance is therefore reduced to choosing how many descriptors to include. Degen et al. (2020) derive redundant modification from the same trade-off between recoverability and cost, with the noise placed in the semantics of the descriptors rather than in the partner. Each descriptor is transmitted through the channel of Equation 6 independently, so an utterance $u$ that uses $k$ descriptors for the message $m$ has the channel probability

$$P(u \mid m; \Pi, \Phi) = \prod_{i=1}^{k} P(u_i \mid m; \Pi, \Phi), \tag{13}$$

where $u_i$ is the $i$th descriptor. In plain terms, every added descriptor supplies the partner with one more independent observation of the message. The illustrations place each descriptor at the message’s canonical form $c_{\Pi}(m)$, the noiseless best case. Producing $k$ descriptors costs $\kappa$ per descriptor. The implementation supplies one candidate utterance per length, so the utterance prior over the candidates is geometric in $k$,

$$P(u) \propto e^{-\kappa k}, \tag{14}$$

and its surprisal is the length cost itself up to an additive constant, $s(u) = \kappa k + \text{constant}$. Choosing $k$ is then an instance of Equation 11. The illustrations set $\kappa$ to 0.05 per descriptor and hold $P_{\text{typ}}(u; \Pi)$ uniform over the candidates, so the utterance prior is fixed across partners.

Production as stated in Equations 10 and 11 corresponds to the speaker layer of the RSA model (Bergen & Goodman, 2015; Goodman & Frank, 2016). In RSA, a cooperative speaker imagines a listener and chooses utterances that lead that listener to the intended message while avoiding needless effort, choosing each utterance with a probability growing with its usefulness:

$$P_S(u \mid m) \propto \exp(\alpha[\log P_{L_0}(m \mid u) - C(u)]). \tag{15}$$

The first quantity, $\log P_{L_0}(m \mid u)$, is how well an imagined listener $L_0$ would recover the message $m$ from the utterance $u$. The second, $C(u)$, is the cost to say it. Their difference is the utterance’s *usefulness*, and the exp converts it into a probability of being chosen. The rationality parameter $\alpha$ sets how strongly the speaker favors the most useful option. Our production mode is this speaker restated. The recovery term is our recovery log-probability $\log P(m \mid u; \Pi, \Phi, \Lambda)$, whose negation is the misrecovery cost of (10). The cost term is our production cost, $C(u) = -\log P(u; \Pi)$. Once $L_0$ is replaced by our noisy-channel listener, the argmax choice in (11) is the limit of (15) as $\alpha$ grows without bound.

Our version differs from the standard RSA speaker in two respects. First, the listener the speaker imagines is not RSA's literal listener but a noisy-channel listener whose fidelity can be high or low. The literal listener takes the utterance at face value and reads it against a prior over world states. That substitution is not itself new: Bergen and Goodman (2015; see Goodman & Frank, 2016) define a literal listener that already decodes through a noise process, and derive a speaker who trades the cost of a fuller utterance against the risk that a shorter one is misrecovered. What is new is that the noise is indexed to a particular partner rather than to the physical channel, so that "speak more explicitly to someone who understands less" follows for the same reason that "speak more explicitly in a noisy room" does. Second, that imagined listener is the same distribution an agent uses in comprehension (Equation 2).

The mirror also settles where the partner's *other* attributes act in production. Speakers design not only *how* they say something, but *what* they choose to say, a property Sacks et al. (1974) named *recipient design* and applied to the selection of topics as well as of words. People are argued on general grounds to design utterances for the particular partner they are addressing (Clark & Murphy, 1982). Speakers shift pronunciation by addressee, as in the studies Bell (1984) reviews. In one study speakers scaled the identifying detail of a referent to how recognizable it was likely to be for listeners of the addressee's kind (Fussell & Krauss, 1992), and in another they mentioned fewer of the events an addressee had already heard (Galati & Brennan, 2010). In the RI model, pronunciation shift comes from the utterance prior $P(u; \Pi)$. The scaling of detail comes from the recovery term $s(m \mid u; \Pi, \Phi, \Lambda)$, with the message prior $P(m; \Pi, \Lambda)$ inside it as the same distribution that told the comprehending agent what this partner was likely to mean. Using it there assumes that the partner as a decoder expects to be told roughly what the agent expects to hear from them. That term weighs the intended message against the other readings available to the partner, so an unexpected message must be spelled out distinctively enough to keep the partner from settling on a more expected reading. The omission of events already heard belongs instead to the choice of what to mean, a step outside Equations 10 and 11.

One step we do not represent is that the agent settles on what to mean before settling on how to say it, and the same partner model should govern that earlier step. Raising a given message with a given partner costs its surprisal under the message prior, $s(m; \Pi, \Lambda) = -\log P(m; \Pi, \Lambda)$: high for a topic unexpected or inappropriate for them, low for one that is not. Pregnancy is an unlikely topic to raise with a man for the same reasons that make "I am *pregnant*" an unlikely thing to attribute to him (Wu & Cai, 2026b). The cost trades off against whatever the agent wants to achieve by speaking, which our objective does not represent. The narrower point stands: choosing what to mean and recovering what was meant would draw on one distribution, just as choosing an utterance and interpreting an utterance do, so the RI model extends in principle to a stage we do not explicitly model here.

Channel fidelity plays the mirror-image role on the production side, read now in the decoding direction. Where the partner is believed to understand well, many utterances will convey the message, so the short default utterance will serve and an agent asking a competent adult partner for a cup needs little more than "the cup". Where the partner is believed to understand poorly, most utterances leave the message likely to be missed, and only fuller ones keep the misrecovery cost low. To a young child or to a voice assistant prone to mishearing, the same agent might say "the red cup on the table". That is what leads an agent to a full noun phrase rather than a pronoun, a carefully articulated vowel rather than a reduced one, or the partner's own word rather than the agent's own.

### 4.6 One model, two goals: symmetry and its limits

The two modes can now be set side by side (Table 2). Each agent holds one variable fixed and chooses the other, scoring candidates by a channel term and a prior over the chosen variable, which draw on the same three parameters between them.

**Table 2.** *The two modes as mirror images of one computation.*

| Aspect | Comprehension | Production |
|---|---|---|
| Agent | Listener | Speaker |
| Partner | Speaker (encoder) | Listener (decoder) |
| Quantity held fixed | Utterance ($u$) | Message ($m$) |
| Quantity chosen | Message ($m^*$) | Utterance ($u^*$) |
| Channel term ($\Pi$ and $\Phi$; also $\Lambda$ in production) | How well a candidate message matches the utterance received | Probability the partner recovers the message from a candidate utterance |
| Prior term ($\Pi$ and $\Lambda$ in comprehension; $\Pi$ in production) | Probability of a message for this partner | Appropriateness of an utterance for this partner |

*Note.* The message prior also acts in production, though only inside the channel term: the partner's recovery of a message already weighs it against what they expected to be told (Section 4.5).

The two modes differ in three respects: the quantity held fixed, the direction in which the channel is read, and the range of the prior term. The RI model is therefore unified across comprehension and production: we posit not two models that share a vocabulary but one model run in two directions. In dialogue the agent alternates between the two modes, reading the same partner model in one direction and then the other. What the agent hears from the partner updates that model, and whether the agent's own messages were understood should update it as well (Section 5.1).

The model now explains why the two adjustments with which we began run in opposite directions, and it is the fidelity part of perceived linguistic competence that does so. The explanation follows the logic of the noisy channel (Bergen & Goodman, 2015; Gibson et al., 2017), here applied to a fidelity that both modes share. A low-fidelity belief flattens the channel in both directions (the shared-fidelity assumption, stated below). In comprehension, the utterance of a low-fidelity partner becomes a weak guide, so the agent falls back on the message prior and appears to *discount* the utterance. In production, most utterances become unlikely to convey the message to the partner, so the agent overrides the utterance prior and appears to *invest* in the utterance. Symmetry of form does not, however, guarantee symmetry of degree. Reliance on the prior carries no cost in the objective of Equation 4 and is available wherever the prior favors some reading, whereas every descriptor a producing agent adds costs effort. Applying the partner model may itself demand processing effort in either mode (Sections 5.2 and 5.3), which the objective does not represent. As fidelity falls, the recoverability added by a further descriptor shrinks and eventually drops below the cost of producing it. We therefore expect discounting to grow steadily as fidelity falls, but investment to grow only to a point before declining.

Figure 3 shows both claims in the implementation of Equation 6, with the message prior of Equation 7 carrying the comprehension case. The comprehension curve is the probability that the agent recovers the plausible non-literal message $m_{\text{alt}}$ rather than the literal reading $m_{\text{lit}}$ on hearing the utterance $u$, the recovery of Equation 12 restricted to the two candidates,

$$P(m_{\text{alt}} \mid u; \Pi, \Phi, \Lambda) = \frac{P(u \mid m_{\text{alt}}; \Pi, \Phi)\, P(m_{\text{alt}}; \Pi, \Lambda)}{P(u \mid m_{\text{alt}}; \Pi, \Phi)\, P(m_{\text{alt}}; \Pi, \Lambda) + P(u \mid m_{\text{lit}}; \Pi, \Phi)\, P(m_{\text{lit}}; \Pi, \Lambda)}. \tag{16}$$

In plain terms, Equation 16 is how likely the agent is to read the utterance as the non-literal alternative rather than literally. As fidelity falls the channel terms equalize and the channel log-likelihood ratio of Equation 8 goes to zero, so the probability rises toward a ceiling set by the message prior alone, $P(m_{\text{alt}};\Pi,\Lambda)/\left[P(m_{\text{alt}};\Pi,\Lambda)+P(m_{\text{lit}};\Pi,\Lambda)\right]$. The production curve is the number of descriptors in the chosen utterance, the choice of Equation 11 over the candidate lengths of Equations 13 and 14,

$$k^{*}=\arg\max_{k}\left[\log P(m\mid u_k;\Pi,\Phi,\Lambda)-\kappa k\right], \tag{17}$$

where $u_k$ is the candidate utterance with $k$ descriptors and the chosen utterance is $u^{*}=u_{k^{*}}$. In plain terms, the agent keeps adding descriptors while each one adds more recoverability than it costs.

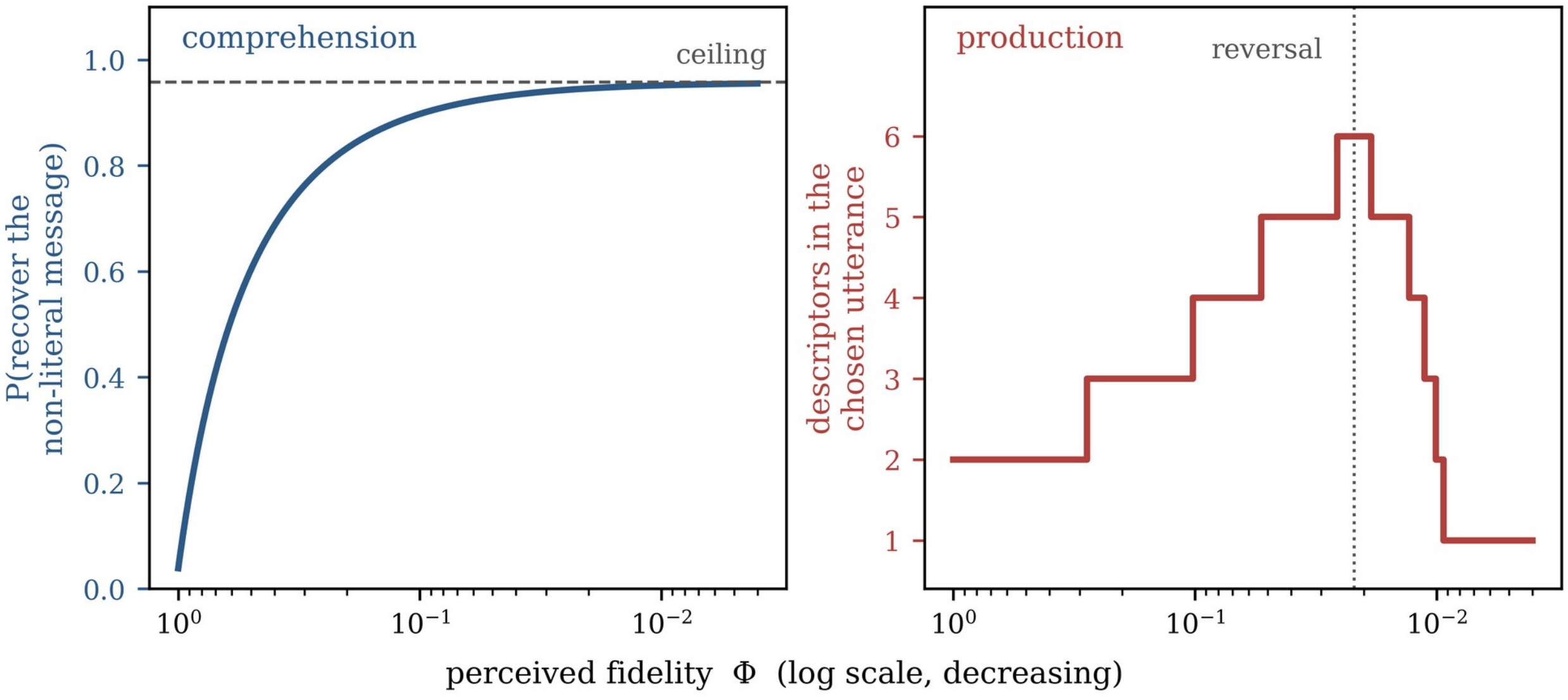

**Figure 3.** *Predicted comprehension and production adjustments over perceived fidelity.* Left panel: the agent's probability of recovering the plausible non-literal message in comprehension, from Equation 16; the dashed line is the ceiling set by the message prior's odds. Right panel: the number of descriptors in the agent's chosen utterance in production, from Equation 17; the dotted line marks the reversal. Comprehension: one prior component at the alternative; $m_{\text{lit}}=0$; $m_{\text{alt}}=1.6$; the heard utterance $u=0$; prior $\sigma=0.60$. Production: two equally weighted prior components at 0 and 0.9; prior $\sigma=0.45$; invariant to $\Lambda$. Both: $\Lambda=0.85$; channel $\sigma=0.45$; $\kappa=0.05$ per descriptor; $k$ searched to 40; $\Phi$ swept from 1.0 to 0.004; recovery normalized over the two candidates; the space $[-8,8]$.

Only the production curve reverses direction. Adding descriptors is worthwhile only while each adds recoverability worth its cost: once Equation 6's unfaithful component dominates, a further descriptor adds almost no discriminating evidence, so the optimum collapses to the shortest utterance. Where the reversal falls depends on the cost per descriptor and on the confusability of the alternatives. How high the peak rises is set by the cost per descriptor alone. Where a descriptor costs more than the gain in recoverability it produces, the model produces no investment, and at costs high relative to the referents' confusability the curve can start at its peak and only fall.

In the RI model a single fidelity governs the partner's channel in both directions, so that fidelity as an encoder and as a decoder are the same. We call this the *shared-fidelity assumption*, and every profile and prediction in this paper uses it; the *transfer* prediction is its test (see Section 6.2). A version with separate encoding and decoding fidelities is possible in principle. Artificial partners are where it would matter most, since a system whose utterances are fluent may still be expected to mishear.

**4.7 Profiles of three partner types: L2 adults, children, and artificial partners**

With both modes in place, we return to the partner types of Sections 2 and 3. The literature reviewed there commonly represents perceived linguistic competence as a single scale. On that scale a fluent native adult ranks above non-native speakers, children, and artificial partners alike. We conjecture that the scale bundles two of the model's three quantities, and that the two dissociate. The first is the fidelity $\Phi$: how reliably a partner is believed to bind forms to meanings. The second is the knowledge $\Lambda$: how knowledgeable the partner is believed to be. The two enter different terms of the computation (Equation 8). The identity $\Pi$ stands outside the bundle: it sets the messages expected of this partner and the forms they use, not how competent they are held to be.

There are three populations that have commonly been studied as linguistically less competent partners in this literature: L2 adults, children, and artificial partners. We hold that they are not one type at three levels but three different types. Those grouped together as "linguistically less competent" occupy different regions of a two-dimensional space of perceived fidelity and knowledge. A single scale predicts that they should behave alike, only to different degrees; two independent quantities predict that they should sometimes behave in opposite ways.

Each type takes its own pair of values for $\Phi$ and $\Lambda$, with one $\Pi$ shared across the profiles. The profiles are schematic. Children and L2 adults are broad populations, a three-year-old differing from a twelve-year-old and a beginner from a near-native learner, so each profile stands for the partners of the studies reviewed in Sections 2 and 3. L2 adults are believed to know the world as well as native adults while binding forms to meanings less reliably: a lowered $\Phi$ with a high $\Lambda$. The profile describes learners of limited proficiency; highly proficient L2 users or balanced bilinguals would lie near the native profile. Children are believed to know less about the world than native adults and to bind forms to meanings less reliably: a lowered $\Phi$ with a lowered $\Lambda$. The children in the production studies reviewed were young, between one and six years old (Cai et al., 2021; Tal et al., 2023), and older children would lie nearer the native profile. Artificial partners invert the L2 profile, with form assumed nearly flawless and content distrusted: a high $\Phi$ with a low $\Lambda$, the knowledge estimate resting on the widely publicized unreliability of language models about facts (Rao et al., 2025a).

The L2 profile is the best supported, combining intact knowledge of the world with a less reliable binding of forms to meanings. The low fidelity attenuates the response to every form violation (Figure 4), and the identity's part in the channel reduces it further for the errors such a speaker typically makes. This would explain why an anticipated agreement error elicited no repair response (Hanulíková et al., 2012), and why the reduction was specific to the errors the speaker's background produces, the typical error losing the late response while an atypical one retained it in delayed form (Caffarra & Martin, 2019). An error typical of that background lies close to the forms such a partner is believed to use, so it carries a small mismatch cost, whereas an atypical one does not. Because their knowledge of the world is believed to be intact, the *content* of a non-native speaker's message need not be discounted. In Lorenzoni et al.'s (2022) studies, Italian participants read a short biography of a native speaker and of a foreign speaker and then rated statements attributed to each. The statements were presented in writing, so the two speakers differed in their described origin rather than in how they sounded. Statements whose truth the participants could not know were rated as making more sense and as more likely to be true when attributed to the foreign speaker, as with "butterflies do not see gray". The leniency was explained by a range of knowledge attributed to a foreigner that differs from the listener's own, which is what a partner-specific $P_{\text{typ}}$ represents.

The child profile lowers both $\Phi$ and $\Lambda$. The lowered fidelity rests mainly on the production evidence. Speakers reuse a young child's labels more than a native adult's (Cai et al., 2021), adopt a child partner's perspective on a referent more often (Y. Zhang & Cai, 2026), and refer more explicitly the younger the child (Tal et al., 2023). In the RI model each of these is investment, the response of a producing agent to a lowered fidelity. In comprehension, the lowered fidelity predicts that comprehenders attribute less pragmatic reasoning to a child speaker. Mayn et al. (2025) asked adults to interpret ambiguous messages in a reference game in which the message was said to come either from another adult or from a four-year-old child. A message such as *red* is ambiguous on its face when a red square, a red triangle, and a third object are in view. Because *square* was not among the available messages while *triangle* was, a speaker meaning the triangle could have said so, and a listener who assumes a reasoning speaker therefore chooses the square. Listeners divided points among the three objects to show their reading of the message, and they assigned the square fewer points for a message attributed to the child than for one attributed to an adult. This suggests that listeners hold a child speaker to a weaker standard of reasoning over alternatives, though those authors also reported substantial variability between individuals. The lowered $\Lambda$ rests on what is believed about a young child's knowledge of the world rather than on a reviewed finding. A remark that is implausible from any speaker has not been compared between a child voice and an adult voice in the studies reviewed. In our model, the child profile's reduced content response is therefore a prediction: such a remark should tend to surprise a comprehender less when it comes from a young child than from a native adult. One further finding about child speakers reflects the identity rather than either quantity. We read the larger cost of a child's switch to a dispreferred synonym (Wu et al., 2024) as an identity effect. A child is expected to be less flexible in word use than an adult and so to keep to the label already established. The effect is of the "bonnet" kind, the identity setting the form expected of the partner. A child's lower $\Phi$ works against it, since a flatter channel charges any departure less, so the identity difference must outweigh the difference in fidelity.

The artificial profile pairs a high $\Phi$ with a low $\Lambda$, and its clearest test comes from comprehending language attributed to a language model. Rao et al. (2025a) recorded brain potentials while participants read sentences attributed either to a large language model or to a human. Under attribution to the language model, an implausible word elicited a smaller brain response and a larger late response. A grammatical error likewise elicited a larger late response under that attribution and no reliable late response under attribution to a human. We read the smaller response to implausible content as the reduced content response of a partner held loosely to what a knowledgeable speaker would mean (Equation 19). The larger response to the grammatical error is the enhanced form response of a partner whose forms are trusted (Equation 18; Figure 4). This suggests that readers expected implausible content from a language model yet corrected its errors more readily. The same low estimate of an artificial partner's knowledge carries the profile's pragmatic prediction. Rao et al. (2025c) presented identical contextually incongruent remarks as coming from an artificial companion or from a human and measured how readily comprehenders read them as ironic. Irony remained the most common reading whatever the source, but comprehenders ascribed a remark to irony less readily when the source was artificial. A low $\Lambda$ predicts this pattern, since an incongruity from a partner who may simply not know is explained at lower cost as error than as an intended message. The authors' own reading, a partly withheld intentional stance, is compatible with that account. A pun attributed to an artificial partner likewise elicited a smaller brain response than one attributed to a human, relative to a control in which the same punchline carried its ordinary idiomatic meaning (Rao et al., 2025b).

The artificial profile is the one most likely to become dated. The early computer-alignment effects rested on a low estimate of linguistic competence, formed against systems assumed to be unsophisticated (Pearson et al., 2006). A system presented as basic rather than advanced is a low-fidelity partner within the artificial type, so alignment with it does not test the artificial profile of Figure 4, which is set for a language model. Alignment appears no longer to follow that estimate, whether toward actual contemporary models (Blevins, 2026) or toward a fluent partner merely believed to be artificial (Li, 2025). The low knowledge estimate should likewise rise as models improve and beliefs follow. Figure 4 marks the consequence with a semitransparent bar, drawing the artificial profile's content response at a contemporary knowledge estimate; that is a prediction rather than a finding. As the knowledge of artificial partners comes to be estimated more highly, the reduced response to content anomalies attributed to them should diminish. On an estimate that has a language model knowing more than a native adult, the reduction should reverse.

The two quantities $\Phi$ and $\Lambda$ appear sufficient to account for the pragmatic findings reviewed here. A pragmatic inference rests on the assumption that the partner chose this form rather than an available alternative, and for a reason. Lowering $\Phi$ undermines that assumption directly. When the mapping from message to utterance is noisy, a modifier or a filled pause or a scalar expression is a poorer index of what the partner decided. The inference then has less evidence to draw on. This would explain why contrastive inference was eliminated for a partner described as impaired and heard over-modifying (Grodner & Sedivy, 2011). It would likewise explain why a filled pause stopped cueing the harder-to-retrieve referent in accented speech (Bosker et al., 2014), and why under-informativeness was forgiven for a non-native speaker with a strong accent (Fairchild & Papafragou, 2018). Lowering $\Lambda$ accounts for the remainder, letting an incongruity be explained without attributing an intention to it, as in the case of a remark's source being artificial (Rao et al., 2025c). Together they set how far the form the partner produced can be trusted to have been chosen deliberately.

Where the two quantities order the three types differently, a two-quantity account disagrees with a single-scale account. Figure 4 shows the predicted costs of a form violation and of a content anomaly. The form response is indexed by the extra surprisal of an ill-formed utterance over its well-formed counterpart at the same message,

$$R_{\text{form}}(\Pi, \Phi) = s(u_{\text{ill}} \mid m; \Pi, \Phi) - s(u_{\text{well}} \mid m; \Pi, \Phi), \tag{18}$$

which depends on $\Phi$ once $\Pi$ is fixed. The content response is indexed by the extra surprisal of an anomalous message over a typical one,

$$R_{\text{content}}(\Pi, \Lambda) = s(m_{\text{anom}}; \Pi, \Lambda) - s(m_{\text{typ}}; \Pi, \Lambda), \tag{19}$$

which depends on $\Lambda$ alone once $\Pi$ is fixed, and the implementation fixes one $\Pi$ across the profiles. Figure 4 shows each response as a ratio to the native adult's value, so the native adult is at one by construction. On these measures L2 adults show a reduced form response alongside a preserved content response. Artificial partners show an enhanced form response alongside a reduced content one. Children show both responses reduced.

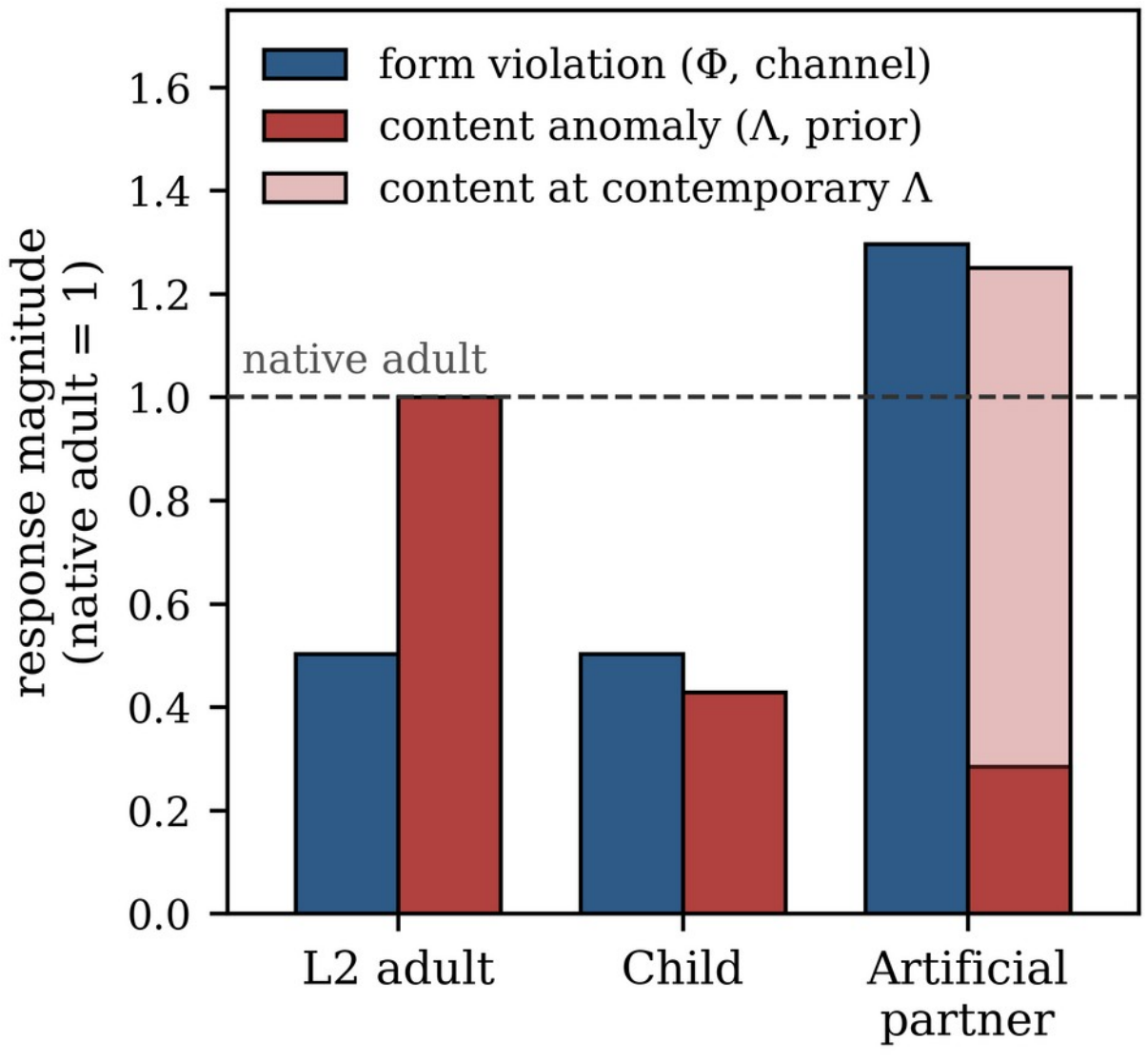

**Figure 4.** *Predicted responses to form violations and content anomalies across the three profiles.* The form response of Equation 18 and the content response of Equation 19 for the three partner types, each relative to the native adult's value (dashed line at one). The semitransparent

bar is the artificial profile's content response at a contemporary knowledge estimate, a prediction rather than a finding. Native adult $\Phi=0.85$ and $\Lambda=0.85$; L2 adult $\Phi=0.35$ and $\Lambda=0.85$; child $\Phi=0.35$ and $\Lambda=0.40$; artificial partner $\Phi=0.97$ and $\Lambda=0.25$; contemporary knowledge estimate $\Lambda=0.93$, at which the content response lies just below the artificial form response. All values are model output at assumed parameter settings, and the profiles are schematic (Section 4.7). $P_{\text{typ}}$: three equal components at 0 and $\pm 2.5$ with $\sigma=0.45$; one $\Pi$ for every profile; $m_{\text{typ}}=0$; $m_{\text{anom}}=5$; $u_{\text{ill}}$ displaced 1.6 from $u_{\text{well}}$; channel $\sigma=0.45$; the space $[-8,8]$.

# 5. Acquiring and using the partner model

The account so far has started from the partner model as given, and as already in use. Three questions remain. The first is how agents arrive at such a model of their conversational partner. The second is when they use it, and the third is to what extent different agents rely on it.

## 5.1 Acquiring and updating a partner model

A partner model is built from an initial stereotype and then updated. On the integrative speaker-model account (Wu & Cai, 2026a; Wu et al., 2025), a comprehender starts from a demographic model of the partner's population and updates it as the partner's own speech and messages accumulate, refining it toward an individualized representation; the stereotype itself is also revised in the process. The ideal-adapter framework describes the same updating for phonetic categories, with a structured prior that allows generalization from a speaker to others of the same kind (Kleinschmidt & Jaeger, 2015). The hierarchical account of convention describes it for the meanings of words (Hawkins et al., 2023). We hold the three parameters at point values, so the RI model itself is static; the updating described in this section is inherited from these accounts. Three routes of acquisition can be distinguished in the evidence.

The default route is by category. An accent establishes a category-level belief before any individuating evidence arrives, as does a child's voice or a label identifying a speaker as artificial. This belief is the demographic model of the integrative account. The categories that carry such beliefs are coarse. A voice heard as male rather than female reversed which contents were expected (Lattner & Friederici, 2003; Van Berkum et al., 2008), as did a voice heard as a child's rather than an adult's or as higher rather than lower in social status (Van Berkum et al., 2008). An accent heard as American rather than British selected which meaning of an ambiguous word was accessed (Cai et al., 2017). A description can suffice: a start-up screen presenting a computer partner as basic produced more alignment than one presenting it as advanced (Pearson et al., 2006).

A second route is by individuating evidence. A speaker's pragmatic reliability can be inferred from their behavior within a session (Gardner et al., 2021), and listeners learned how a recently encountered speaker used particular words, for instance how far along a probability scale that speaker's "might" reached (Schuster & Degen, 2020). That learning is speaker-specific when the same listener hears two speakers: listeners came to expect each speaker's quantifier use separately after exposure to one speaker biased toward "some" and another biased toward "many" (Yildirim et al., 2016). It is not confined to word meaning, since comprehenders also acquired a speaker's syntactic preferences and carried the learned bias over to sentences containing a verb and theme not heard during exposure (Kamide, 2012). Listeners also track how often an individual speaker violates stereotypes: the tracking was reflected in neural oscillations, partly general across speakers and partly specific to the individual speaker (Wu et al., 2025).

A third route is by the partner's record of accuracy, which informs the knowledge parameter $\Lambda$. The developmental evidence indicates that the computation and its updating are in place early. Three-year-olds registered which speaker had been inaccurate, and from around age four they used that record to decide whom to believe (Koenig & Harris, 2005). Four- and five-year-olds, though not yet three-year-olds, allowed demonstrated accuracy to override accent when the two conflicted (Corriveau et al., 2013). Where no accuracy record was available, accent alone still guided learning beyond language, since children of those ages endorsed a native-accented speaker's silent demonstration of an object's function over a foreign-accented speaker's (Kinzler et al., 2011). Evidence of communicative success, whether the agent's own messages were understood, would supply a further route beyond these three; speakers who had been misunderstood hyperarticulated the misheard contrast in their later productions (Buz et al., 2016).

Children in this age range also apply the computation to their own partners. Rett and White (2022) played five- and six-year-olds recordings of two women labeling familiar objects, one with correct subject-verb agreement ("that *is* a duck") and one without ("that *am* a ball"). The two speakers were native for one group of children and foreign-accented for another; each speaker then applied the same new word to a different unfamiliar object, and the children were asked which object the word named. Children who had heard the native speakers mostly chose the grammatical speaker's object, whereas those who had heard the same errors in foreign-accented speech chose at chance. Both groups nonetheless identified the ungrammatical speaker as the one who had “said silly things”, with no reliable difference between them. This suggests that children detected the errors from either speaker but gave them less weight in judging a foreign-accented speaker’s reliability. On the production side, Shatz and Gelman (1973) reported that four-year-olds explaining how a toy works simplified their speech for a two-year-old; in addition, their speech to same-age peers resembled their speech to adults. The adjustment therefore appears to track the young listener’s capacities rather than a simple adult/child division.

### 5.2 When the partner model is used

Holding a partner model does not mean consulting it in every situation. The findings reviewed suggest four conditions under which it is consulted. The first condition is that the belief about the partner is clearly established. Grodner and Sedivy’s (2011) unreliable speaker was described as impaired and also behaved erratically. Gibson et al.’s (2017) speaker had a strong foreign accent. Cai et al.’s (2021) partner spoke in a child’s voice. Pearson et al.’s (2006) computer was labeled as a basic system. When the description of impairment was withheld and the speaker merely over-modified, the speaker effect disappeared (Grodner & Sedivy, 2011). The second condition is that the utterance leaves competing readings close before the partner is taken into account. A partner belief rescales one term of Equation 8, so a moderate change of belief can change which reading wins only where the two terms are of comparable size. That was the case for an ambiguous “ferry” against a display themed on imaginary creatures (Lev-Ari, 2015), for an implausible literal reading (Gibson et al., 2017), and for an under-informative statement (Fairchild & Papafragou, 2018). Whether the readings are close is a property of the utterance and its candidates rather than of the belief about the partner, so the condition can be assessed before that belief is applied. The third condition is that the evidence departs from how speakers ordinarily behave. Listeners generalized from a speaker who under-modified but not from one who over-modified, and over-modification is common in ordinary speech (Pogue et al., 2016). The fourth condition is that the agent has processing resources to spend. The anticipatory shift toward the theme-consistent object grew with working-memory span (Lev-Ari, 2015), a result that we also interpret as a trait difference in Section 5.3.

### 5.3 To what extent the partner model is used

Everything the partner model has contributed so far has been a belief about the partner, but two agents holding the same belief need not rely on it to the same degree. One quantity therefore has to be added: how much weight the agent places on the partner model.

We write $\omega$ for that weight, with $0 \leq \omega \leq 1$. The three primitive partner-dependent distributions are the channel and the message prior of Equation 2 and the utterance prior of Equation 11. Each is blended with the agent's generic expectation:

$$P_\omega(m; \Pi, \Lambda) = \omega P(m; \Pi, \Lambda) + (1 - \omega) P(m), \tag{20}$$

$$P_\omega(u \mid m; \Pi, \Phi) = \omega P(u \mid m; \Pi, \Phi) + (1 - \omega) P(u \mid m), \tag{21}$$

$$P_\omega(u; \Pi) = \omega P(u; \Pi) + (1 - \omega) P(u). \tag{22}$$

In plain terms, $\omega$ is the degree to which the agent takes the particular partner into account. At $\omega = 1$ the account is exactly as stated in Section 4; at $\omega = 0$ the agent comprehends and produces with generic expectations, and no manipulation of the partner should have any effect. $\Pi$, $\Phi$, and $\Lambda$ are beliefs the agent holds *about* the partner; $\omega$ is instead a property *of the agent*.

Across partners, $\omega$ is distinguishable from the partner parameters because it scales every partner effect at once, whichever of the three parameters the effect comes from. A change in $\Phi$ instead moves only the effects that run through the channel. Within a single partner the two are not distinguishable, because the blend of Equation 21 is equivalent to shifting $\Phi$ toward the generic value when the two channels share their canonical forms, as they do in the implementation.

Figure 5 shows the partner-weight pattern in both modes. Each panel follows two partners, one high-fidelity and one low-fidelity, as $\omega$ runs from 0 to 1. The curves are computed from Equations 16 and 17, with the blended channel of Equation 21 in place of the partner-conditioned one. The blends of Equations 20 and 22 have no effect here, the partners sharing one message prior and the implementation's utterance prior being partner-free. The generic expectation is anchored at the native-adult fidelity, reading no partner information as the assumption of a typical native adult, so the high-fidelity partner's curves stay flat and the low-fidelity partner carries the whole effect. At $\omega = 0$ the two partners produce identical behavior, as they must. As $\omega$ rises, the comprehension curve for the low-fidelity partner departs smoothly and without reversal. The production curves instead remain identical until the weight exceeds a threshold, after which the low-fidelity partner receives more descriptors in steps of one.

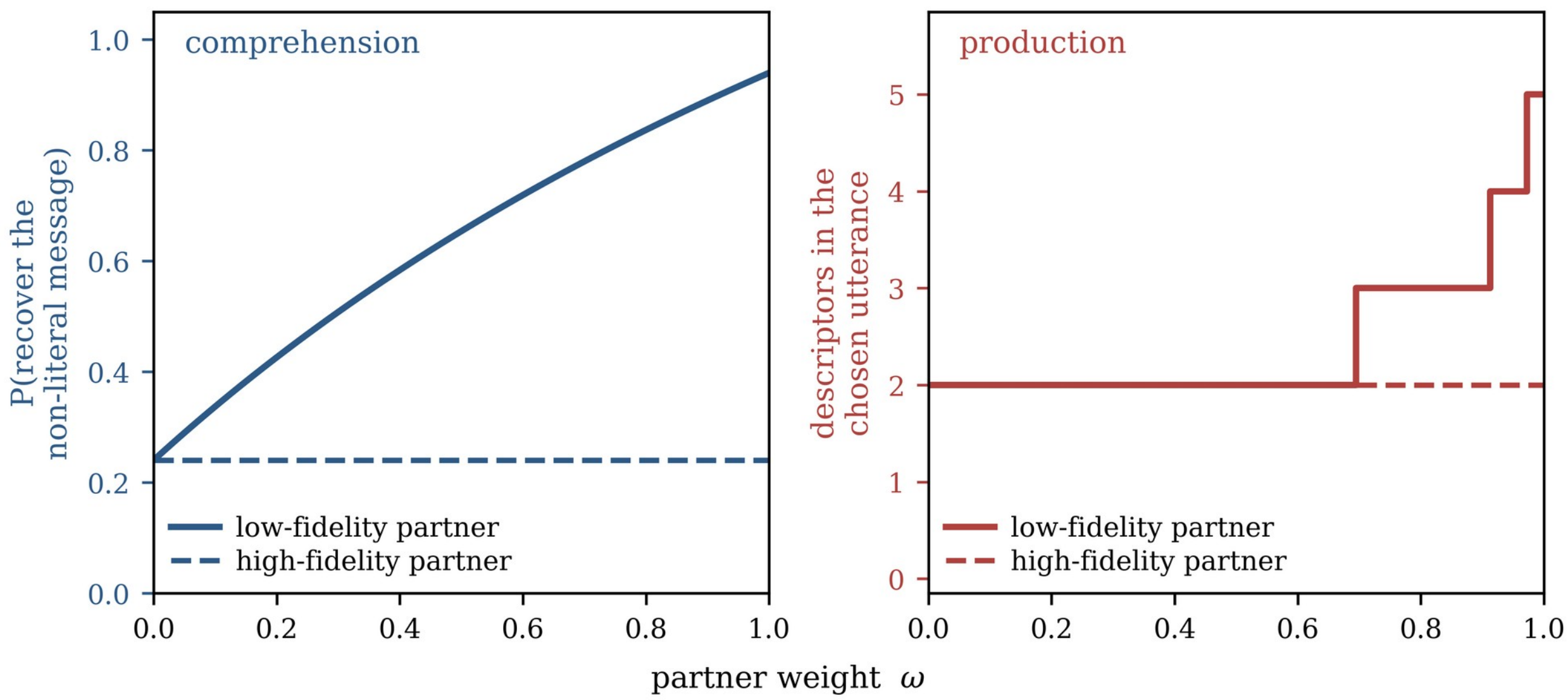

**Figure 5.** *Predicted comprehension and production adjustments over partner weight.* Left panel: the probability of recovering the plausible non-literal message (Equation 16) for a low-fidelity and a high-fidelity partner, plotted against the partner weight $\omega$. Right panel: the number of descriptors in the chosen utterance (Equation 17) for the same two partners. The two partners differ only in the channel of Equation 21; Equations 20 and 22 have no effect, the partners sharing one message prior and the utterance prior being partner-free. Settings are those of Figure 3, except $\Phi_{low} = 0.03$ and $\Phi_{high} = 0.85$; the generic expectation is set at the native-adult value $\Phi = 0.85$.

Individual differences in $\omega$ appear in both comprehension and production. $\omega$ is our formulation of the agent-side properties that determine to what extent the partner is taken into account, so a trait that scales the partner's influence on comprehension or production is consistent with a difference in it. On the production side, speakers lower in working memory and executive control produced more references that took account of what only they could see rather than what their addressee could see (Wardlow, 2013). On the comprehension side, the online adjustment toward the theme-consistent object was graded by working memory (Lev-Ari, 2015). Agents higher in self-reported empathy showed a larger N400 difference between messages that fit a partner's voice-based identity and messages that did not (van den Brink et al., 2012). Agents higher in openness showed a smaller N400 to content that violated a social stereotype for the speaker's voice (Wu & Cai, 2026b).

Autism is a typical candidate for a difference in $\omega$. Autistic individuals often encounter challenges in social and language communication (Loveland & Landry, 1986; Mundy et al., 1986; Schaeffer et al., 2023), and in particular they tend to be less sensitive to the social and emotional information during language communication (Kausel et al., 2024; Lartseva et al., 2015; Wu et al., 2026). The findings that matter here are those in which the social information is about the partner. Autistic and non-autistic adolescents in Groen et al. (2010) heard speaker-content mismatches: sentences whose content did not fit the sex, age, or social class suggested by the voice. The autistic adolescents showed reduced activation of the left inferior frontal region for these mismatches. Autistic adults given the same kind of mismatch detected it as well as non-autistic adults did when asked afterward, but they showed additional activation of the right inferior frontal region while listening (Tesink et al., 2009). The pattern was read as social knowledge that is intact but accessed less automatically.

The cues to a speaker's intent show the same pattern. Wang et al. (2006) asked autistic children to decide whether the speaker's final remark in a short scenario was sincere or ironic; the autistic children were less accurate than non-autistic children, particularly where the cue lay in the context. In a related study by Wang et al. (2007), autistic children showed reduced activity in the medial prefrontal cortex and the right superior temporal gyrus during such remarks. The medial prefrontal activity rose in the autistic group only when they were instructed to attend to the speaker's facial expression and tone of voice. The reduced sensitivity extends to the voice itself: autistic adults rated emotional prosody as less intense than non-autistic adults did (Gebauer et al., 2014), and their voice-sensitive cortex responded typically to voices but atypically when the task was to recognize whose voice it was (Schelinski et al., 2016).

We read the findings on speaker identity and intent as a lower $\omega$. The partner's identity or intent was available, since autistic participants detected the speaker mismatches (Tesink et al., 2009) and engaged the medial prefrontal cortex for the social cues once directed to them (Wang et al., 2007), but they weighted that information less when not directed to it.

## 6. Situating and testing the RI model

### 6.1 Relation to previous accounts

The RI model develops a line of work that runs from interlocutor modeling (Cai et al., 2017, 2021). Cai et al. (2017) proposed a speaker-model account of spoken word recognition, on which comprehenders infer key characteristics of their interlocutor from the speech signal and use that information to guide access to word meaning. Cai et al. (2021) further argued that speaker modeling in comprehension and interlocutor-specific audience design in production are one process viewed from the two ends of a conversation. People build a model of their interlocutor and use it to constrain comprehension and to tailor production alike, with the interlocutor's linguistic competence among the attributes represented. The integrative model (Wu & Cai, 2026a) extended interlocutor modeling on the comprehension side. On that model, prior beliefs about a speaker condition comprehension at phonetic, lexical, and semantic levels. The unfolding speech and message continuously update the speaker model, refining the demographic model into an individualized representation.

We carry this line of work a step further by turning it into a computation. The RI model adds four elements to the earlier accounts. First, the claim that one model serves both modes becomes one computation: the same partner model enters the channel of comprehension and the recovery term of production. That computation shows why the two adjustments to a linguistically less competent partner run in opposite directions. Second, the interlocutor model becomes three parameters with fixed roles: the identity $\Pi$ sets the canonical form and the typical messages; the fidelity $\Phi$ weights the channel; and the knowledge $\Lambda$ weights the message prior. Third, linguistic competence is a single attribute in the earlier accounts; we decompose it into fidelity and knowledge, which order the three partner types differently on form and on content responses. Fourth, the partner weight $\omega$ scales how far the partner model is used. We draw four predictions from these additions in Section 6.2.

In formal terms, the RI model joins two existing formalisms. Comprehension is speaker modeling within the noisy-channel model (Gibson et al., 2013; Levy, 2008), and production is audience design built on the speaker layer of the rational speech act model (Frank & Goodman, 2012; Goodman & Frank, 2016). What we add to each is the partner model. In the noisy channel, the noise model and the prior become jointly partner-dependent through three partner parameters ($\Pi$, $\Phi$, and $\Lambda$). The internal listener of the rational speech act model becomes the partner-conditioned recovery term of Equation 12. Relative to both formalisms, what is new is that one partner-indexed fidelity serves both modes and that perceived competence divides into fidelity and knowledge.

Among the other accounts, the memory-based account poses the most direct challenge to any proposal stated in terms of a partner *model*: partner-sensitive behavior may emerge from ordinary memory rather than from a dedicated partner representation. On that account, a partner cues the associations formed with them, and audience-design effects follow from the structure of what has been encoded without a dedicated process of reasoning about the partner's state (Horton & Gerrig, 2005a, 2005b). A direct test found no amplification of the partner effect when speakers had to recall the referent from memory, though the task made it easy to access which partner knew an item (Ahn & Brown-Schmidt, 2020). The account argues against special-purpose representations and processes rather than against speakers representing their partners, and it allows partner information to be recruited both automatically and strategically (Horton & Gerrig, 2016). On a related view the partner is a context enriched with that person's perspective and knowledge rather than a bare cue (Brown-Schmidt et al., 2015). The memory-based account is therefore a hypothesis about the mechanism that implements partner-sensitive behavior, compatible with a computational-level account of what that behavior achieves. Relative to it, the RI model adds a representation the partner-as-cue view does not itself specify: a belief about how faithfully this partner encodes and decodes.

Two rational accounts already hold partner-specific beliefs and learn them hierarchically. The ideal-adapter account of speech perception (Kleinschmidt & Jaeger, 2015) holds that a listener maintains beliefs about speaker-specific generative models and updates them as evidence accumulates. It is worked out for the mapping between phonetic categories and acoustic cues. The hierarchical account of convention (Hawkins et al., 2023) does the same for the meanings of words: speaker and listener hold the same partner-specific mapping from meanings to words, learned from a population-level prior toward the individual partner, and both modes read it. Of the accounts discussed here it is the closest to ours in scope. What both accounts represent about a partner is which mapping the partner uses: from phonetic categories to acoustic cues in the one account and from meanings to words in the other. In the RI model that mapping is the canonical form $c_{\Pi}(m)$ that the identity $\Pi$ fixes for each message (Section 4.2), and $\Pi$ also sets the messages typical of the partner (Section 4.4). The ideal adapter also infers a speaker-specific precision. The RI model adds two graded weights on the mapping, the fidelity $\Phi$ and the knowledge $\Lambda$, and carries them into production.

Work on perspective taking and common ground posits distinct representations of self and other (Brown-Schmidt & Heller, 2018); multiple-perspectives theory adds a process that compares them continuously during conversation (Heller & Brown-Schmidt, 2023). That tradition already attributes knowledge to a partner on the basis of a social category. It describes speakers as selecting forms by whether a partner can decode them, as when a word is spelled out to keep a preliterate child from understanding it (Brown-Schmidt & Heller, 2018). What a partner can and cannot understand can therefore be represented within it. Relative to multiple-perspectives theory, the knowledge $\Lambda$ is a coarse and single-weight version of what that theory represents as the other's knowledge, whereas neither $\Pi$ nor $\Phi$ represents the partner's perspective or what self and other share. Common ground (what both interlocutors know and know that they share) is therefore a different construct from the knowledge $\Lambda$, which is one-sided. Fidelity is orthogonal to both: one can know exactly how knowledgeable a partner is and still model how faithfully they will encode or decode. What we add is not fidelity as a dimension, which that theory could accommodate, but a graded parameter for it.

The best-known integration account of production and comprehension (Pickering & Garrod, 2013) interweaves the two by prediction: speakers predict the perceptual consequences of their own upcoming utterance through a forward model, and listeners covertly imitate the speaker to predict what the speaker will say next. The RI model integrates the two modes at the computational level, through one partner model, and could be implemented by forward models of that kind. The two-route account of audience design (Ferreira, 2019) is specific to production and divides audience design by processing route rather than by content. A feedforward route can act only on information available before production begins, and child-versus-adult addressee status is offered as the leading example of such an already-known addressee property, with the limited proficiency and attention span that status entails. A recurrent route instead runs a forward model over the utterance the speaker is about to produce, and it can also act on idiosyncratic properties of the particular addressee. That forward model generates communicatively relevant features separately from full production and evaluates them through comprehension, and the account represents the addressee's specific properties without parameterizing them. Relative to it, the RI model specifies those properties as the partner's fidelity and knowledge, and makes them the same quantities that govern comprehension. A category-level belief about a partner's fidelity is available before production begins, so it can act on the feedforward route.

### 6.2 Predictions

We now make four predictions, each a direction rather than a magnitude. Any measure of processing cost, interpretive choice, or explicitness can serve as the index.

*The channel-prior dissociation.* We place a partner's fidelity in the channel and not in the message prior. We therefore expect lower perceived fidelity to reduce the cost of a form violation and how far it is repaired, while leaving the cost of a content anomaly comparatively unchanged once the general cost of an accented signal is set aside (as in Hanulíková et al., 2012). An identity manipulation such as regional origin should instead tend to move which form is expected, an effect that should weaken as perceived fidelity falls. After an extended run of well-formed utterances, a non-native speaker's errors should come to be corrected as a native speaker's are.

*The knowledge-fidelity dissociation.* We expect perceived fidelity and perceived knowledge to be manipulable independently and to do different work. Describing a partner as an expert or a novice should mainly affect how content anomalies are processed, whereas presenting the same partner's speech as strongly or barely accented should mainly affect how form violations are processed. The three profiles of Section 4.7 could be fixed independently of the responses they predict by rating the partner types for perceived reliability and knowledge.

*The comprehension ceiling and production reversal.* Both adjustments should have a limit. As a partner's perceived fidelity falls, we expect non-literal interpretation to rise toward a ceiling set by the message prior, so that a manipulation of plausibility moves the ceiling and a manipulation of fidelity does not. In production, we expect the explicitness a speaker supplies to rise as perceived fidelity falls and then to decline once a further addition is no longer worth its cost, an inverted U rather than a monotone increase.

*Transfer across modes.* One fidelity enters the channel of comprehension and the recovery term of production (the shared-fidelity assumption), so a change in perceived fidelity produced in one mode should carry over to the other for the same partner. Perceived fidelity can be changed by exposure (Gibson et al., 2013). Once listeners have learned to discount a particular speaker's utterances, we expect them to invest more when speaking to that speaker. The partner weight $\omega$ estimated in one mode should likewise tend to predict the individual's adjustment in the other. Absence of transfer, with the within-mode effects intact, would favor separate encoding and decoding fidelities that still share the identity and the knowledge.

## 7. Conclusion

We propose the rational interlocutor (RI) model, on which the adjustments that comprehenders and producers make for their partner are one rational policy viewed from the two ends of a conversation. Interlocutors hold a model of their partner. The partner's identity sets what they are expected to mean and the forms they use. Their fidelity sets how reliably messages and utterances are believed to map onto each other for them. Their knowledge sets how likely their messages are to keep to those of a knowledgeable speaker. Comprehension chooses the message that best reconciles the utterance with the expectation. Production chooses the utterance from which the partner is most likely to recover the message, once the effort of saying it is allowed for. A low-fidelity belief flattens the same channel in both modes, which makes a comprehender rely on the expectation and makes a producer add signal. We argue that perceived linguistic competence is not one quantity but two. Fidelity is a belief about how reliably the partner encodes and decodes, and it belongs to the channel. Knowledge is a belief about how knowledgeable the partner is, and it belongs to the message prior. L2 adults, children, and artificial partners combine the two differently. The model therefore predicts that the three partner types will sometimes behave in opposite ways rather than as one type at three levels.

## Funding

This work was supported by the General Research Fund (grant numbers: 14600220 and 14601525), University Grants Committee, Hong Kong.

# Supplementary materials

*For "A model of rational interlocutors: Unification of comprehension and production."*

## S1. Notation conventions

Four conventions govern the symbols of the main text. Lowercase roman letters are the values variables take: $m$ a particular message, $u$ a particular utterance, $k$ the number of descriptors in an utterance and $p$ a probability or a density. The exceptions are the surprisal function $s$ and the canonical-form function $c_{\Pi}$ that maps a message to the form a partner of identity $\Pi$ uses for it. Subscripted forms such as $m_{\text{alt}}$, $m_{\text{lit}}$, $m_{\text{typ}}$, $m_{\text{anom}}$, $u_k$, $u_{\text{well}}$, and $u_{\text{ill}}$ pick out particular candidate messages and utterances. Uppercase roman letters name functions: $P$, $C$, $R$, $N$ and Uniform. The exceptions are $L_0$ and the subscript $S$, which label the literal listener and the speaker of the rational speech act model in Equation 15. Uppercase Greek letters are the agent's beliefs about the partner: $\Pi$, $\Phi$ and $\Lambda$. They are always listed in that order, and each term lists only the parameters it depends on. Lowercase Greek letters are properties of the agent or of the implementation: $\alpha$, $\omega$, $\kappa$ and $\sigma$.

**Table S1.** *The equations by kind, with the values used in the figures.*

| Equation | | Values as implemented |
|---|---|---|
| 1 | Conceptual | None; the unparameterized rule carried into Equation 2 |
| 2 | Conceptual | Simulated only through Equations 6 and 7 |
| 3 | Conceptual | Definition of surprisal; the figures take logarithms to base e, so all costs are in nats |
| 4 and 5 | Conceptual | Comprehension mode; the figures plot the graded recovery of Equation 16 rather than the argmax |
| 6 | Implementation | $\sigma = 0.45$, the space $[-8, 8]$, and the canonical form $c_{\Pi}(m) = m$ for every partner in every figure but the schematic; $\Phi = 0.9$ and $0.4$ with $c_{\Pi}(m)$ at $-0.5$ and $1.3$ (Figure 2); $\Phi = 0.92$ and $0.18$ (Figure 1); swept from 1.0 to 0.004 (Figure 3); profile values (Figure 4); 0.85 and 0.03 against a generic 0.85, the native-adult value (Figure 5) |
| 7 | Implementation | $\Lambda = 0.9$ and $0.4$ with one component at $-0.5$ and $1.3$ (Figure 2); $\Lambda = 0.85$ except Figure 4's profiles; the components of $P_{\text{typ}}$ per figure note |

| **Equation** | | **Values as implemented** |
|---|---|---|
| 8 | Conceptual | Exact identity from Equation 2 |
| 9 | Conceptual | Exact substitution of Equation 7 into the prior log-odds gap of Equation 8 |
| 10 and 11 | Conceptual | Production mode; implemented as the length choice of Equation 17 |
| 12 | Conceptual | Bayes-rule identity for the recovery; normalized over each figure's candidate messages |
| 13 | Implementation | Each factor is the channel of Equation 6, evaluated at the message's canonical form $c_{\Pi}(m) = m$; k searched from 1 to 40 |
| 14 | Implementation | $\kappa = 0.05$ per descriptor in Figures 3 and 5; $P_{typ}(u; \Pi)$ uniform over the candidate lengths, so the utterance prior is partner-free |
| 15 | Conceptual | The RSA speaker for reference; α does not enter the figures |
| 16 | Implementation | Figures 3 and 5: the heard utterance at the literal reading; the non-literal alternative 1.6 away; prior component width 0.60; $\Lambda = 0.85$ |
| 17 | Implementation | Figures 3 and 5: referents 0.9 apart with equal prior weights; $\sigma = 0.45$; $\kappa = 0.05$ |
| 18 and 19 | Implementation | Figure 4: ill-formed utterance displaced 1.6; anomalous message at 5.0 against a typical one at 0; profile $(\Phi, \Lambda)$ pairs in the Figure 4 note |
| 20 and 22 | Conceptual | Partner-weighted blends of the message prior and the utterance prior; inert in Figure 5, which uses one message prior and a partner-free utterance prior |
| 21 | Implementation | Figure 5: $\Phi = 0.03$ and 0.85 against a generic 0.85, the native-adult value, with ω swept from 0 to 1 |

*Note.* A conceptual equation states the account and would survive any reasonable change of implementation; an implementation equation is a definite form the figures compute, and its committed values are listed. Every quantity plotted in Figures 1 to 5 derives from the implementation equations above. Densities are evaluated on a grid of 1,601 points over [−8, 8], fine enough that the plotted values match closed-form evaluation.